\documentclass[journal]{IEEEtran}

\usepackage{ifpdf}
\usepackage{cite}

\usepackage{multirow}
\usepackage{float}
\usepackage{times}
\usepackage{amsmath}
\usepackage{algorithm}
\usepackage{array}
\usepackage{subfigure}
\usepackage{fixltx2e}
\usepackage{stfloats}

\usepackage{xcolor}
\usepackage{amssymb}
\usepackage{amsthm}
\usepackage{url}
\usepackage{algorithmic}
\usepackage{colortbl}
\usepackage[T1]{fontenc}
\usepackage{graphicx}
\usepackage{epstopdf}
\usepackage{slashed}

\usepackage{multirow,booktabs,color,soul,threeparttable}

\usepackage{xr}
\newcommand{\ie}{\mbox{\emph{i.e.,}}}
\newcommand{\eg}{\mbox{\emph{e.g.,}}}
\newcommand{\etal}{\mbox{\emph{et al.}}}
\definecolor{hl}{rgb}{0.75,0.75,0.75}
\sethlcolor{hl}

\makeatletter
\newcommand{\removelatexerror}{\let\@latex@error\@gobble}
\makeatother

\begin{document}

\title{Constrained Multi-objective Optimization with Deep Reinforcement Learning Assisted Operator Selection}

\author{Fei~Ming,
            Wenyin~Gong, \IEEEmembership{Member, IEEE},
            Ling~Wang,~\IEEEmembership{Member, IEEE},
            and Yaochu~Jin, \IEEEmembership{Fellow, IEEE}
\thanks{This work was partly supported by the National Natural Science Foundation of China under Grant Nos. 62076225 and 62073300, and the Natural Science Foundation for Distinguished Young Scholars of Hubei under Grant No. 2019CFA081. \emph{(Corresponding authors: Wenyin Gong \& Yaochu Jin.)}}
\thanks{F. Ming and W. Gong are with the School of Computer Science, China University of Geosciences, Wuhan 430074, China. (Email: wygong@cug.edu.cn)}
\thanks{L. Wang is with the Department of Automation, Tsinghua University, Beijing 100084, China. (Email: wangling@tsinghua.edu.cn)}
\thanks{Yaochu Jin is with the Faculty of Technology, Bielefeld University, North Rhine-Westphalia, 33619, Bielefeld, Germany (Email: yaochu.jin@uni-bielefeld.de).}
}

\markboth{IEEE/CAA JOURNAL OF AUTOMATICA SINICA,~Vol.~X, No.~X, X~X}%
{Shell \MakeLowercase{\textit{et al.}}: Bare Demo of IEEEtran.cls
for Journals}

\maketitle

\begin{abstract}
Solving constrained multi-objective optimization problems with evolutionary algorithms {has attracted} considerable attention. Various constrained multi-objective optimization evolutionary algorithms (CMOEAs) have been developed with the use of different algorithmic strategies, evolutionary operators, and constraint-handling techniques. The performance of CMOEAs may be {heavily dependent on} the operators used, however, it is {usually} difficult to select suitable operators for the problem at hand. Hence, improving operator selection is promising and necessary for CMOEAs. This work proposes an online operator selection framework assisted by Deep Reinforcement Learning. {The dynamics of the population, including convergence, diversity, and feasibility, are} regarded as the {\it state}; the candidate operators are considered as {\it actions}; and the improvement of the population state is treated as the {\it reward}. By using a Q-Network to learn a policy to estimate the Q-values of all actions, the proposed approach can adaptively select an operator that maximizes the improvement of the population according to the current state and thereby improve the algorithmic performance. The framework is embedded into four popular CMOEAs {and assessed on 42 benchmark problems}. The experimental results reveal that {the proposed Deep Reinforcement Learning-assisted operator selection} significantly improves the performance of these CMOEAs and the resulting algorithm obtains better versatility compared to nine state-of-the-art CMOEAs.
\end{abstract}


\IEEEpeerreviewmaketitle

\section{Introduction}\label{sec:intro}

\IEEEPARstart{C}{onstrained} multi-objective optimization problems (CMOPs) {contain} multiple conflicting objective functions and constraints, which widely exist in real-world applications and scientific research~\cite{Kumar2021Benchmark-Suite}. For example, web service location-allocation problems~\cite{Tan2021Evolutionary} and vehicle scheduling of urban bus lines~\cite{Ma2021Shift}.

The development of solving CMOPs by constrained multi-objective optimization evolutionary algorithms (CMOEAs) has seen prominent growth due to the wide existence of this kind of problem. Generally, a CMOEA contains three key components that affect its performance: the algorithmic strategy used to assist the selection procedures, the constraint-handling technique (CHT) to handle constraints, and the evolutionary operator to generate new solutions. In the past decade, a large number of CMOEAs have been proposed, most of which focus on enhancing algorithmic strategies~\cite{Tian2020Coevolutionary,Tian2021Balancing,Jiao2022Multiform,Qiao2022An} and CHTs~\cite{Ma2021Shift,Ma2021New}. {By contrast, little research focused on how to adaptively select the evolutionary operator in CMOEAs has been reported in the literature.}

According to~\cite{Tian2022Principled}, different evolutionary operators~\footnote{An evolutionary operator means the operations of an evolutionary algorithm used for generating offspring solution, such as the crossover and mutation of GA, the differential variation of DE, and the particle swarm update of PSO/CSO.} are suited to different optimization problems. For example, crossovers in Genetic Algorithms (GAs)~\cite{Holland1994Adaptation} are effective in overcoming multimodal features; mutations in Differential Evolution (DE)~\cite{Storn1997Differential} can handle complex linkages of variables because they use the weighted difference between two other solutions. The Particle Swarm Optimizer (PSO)~\cite{Obayyanahatti1995New} shows good performance on convergence speed and the Competitive Swarm Optimizer (CSO)~\cite{Zhang2018competitive} is good at dealing with large-scale optimization problems because of its competitive mechanism, {and the multi-objective PSO (MOPSO)~\cite{Hu2023FCAN-MOPSO,Hu2020Variational} is particularly designed and effective for multi-objective situations}. Given the fact that a real-world CMOP is usually subject to unknown features and challenges, using a fixed evolutionary operator will limit the applicability of a CMOEA. Although {ensemble and adaptive} selection of operators {have received increased attention} in the multi-objective optimization community~\cite{Wang2019Evolutionary,Tian2022Deep,Dong2022Adaptive}, unfortunately, no research efforts have been dedicated to constrained multi-objective optimization. Hence, it is necessary to study the effect of adaptive operator selection in dealing with CMOPs.

Among the Reinforcement Learning community, Deep Reinforcement Learning (DRL) is an emerging topic and has been applied in several real-world applications in the multi-objective optimization field~\cite{Schneider2021Self-Learning,Caviglione2021Deep,Liu2022Hybridization,Li2022Many-Objective,Zhao2023Hyperheuristic}. Also, its effectiveness in dealing with multi-objective optimization problems (MOPs) has been studied recently~\cite{Li2021Deep,Tian2022Deep,Zhang2022Meta-Learning-Based}. Since DRL uses a deep neural network to approximate the action-value function, it can handle continuous state space and thereby is suitable for operator selection for CMOPs because a population could have infinite {states} during the evolution. {Therefore, it is promising to adopt DRL techniques to fill the research gap of operator selection in handling CMOPs. However, when applying DRL to CMOPs, the main challenge is to properly consider constraint satisfaction and feasibility in a DRL model.}

This work proposes a DRL-assisted online operator selection framework for CMOPs. Specifically, the main contributions are as follows:
\begin{enumerate}
 \item We propose a novel DRL model for operator selection in CMOPs. The state of the population, including the convergence, diversity, and feasibility performance, is viewed as the state, the candidate operators are regarded as actions, and the improvement of the population state is the reward. Then, we develop a Deep Q-Learning (DQL)-assisted operator selection framework for CMOEAs. A deep Q-Network (DQN) is trained to learn a policy that estimates the Q-value of {an} action at the current state.
 \item The proposed model can contain {an arbitrary} number of operators and the proposed framework can be easily employed in any CMOEA. In this work, we develop an Operator Selection (OS) method and further {instantiate} it by using GA and DE as candidate operators in the DRL model. Then, the OS is embedded into four popular CMOEAs: CCMO~\cite{Tian2020Coevolutionary}, PPS~\cite{Fan2019Push}, MOEA/D-DAE~\cite{Zhu2020Constrained}, and EMCMO~\cite{Qiao2022An}.
 \item Extensive experimental studies on four popular and challenging CMOP benchmark test suites demonstrated that the OS method can significantly improve the performance of CMOEAs. Furthermore, compared to nine state-of-the-art CMOEAs, our methods obtained better versatility on different problems.
\end{enumerate}

The remainder of this article is organized as follows. Section~\ref{sec:related-work} introduces the related work. Section~\ref{sec:proposed} {elaborates on} the proposed methods. Then, experiments and analysis are presented in Section~\ref{sec:experiments}. Finally, conclusions and future research directions are given in Section~\ref{sec:conclusions}.

\section{Related Work and Motivations}\label{sec:related-work}

{
\subsection{Preliminaries}\label{sec:related-work-preliminary}
Without loss of generality, a CMOP can be formulated as:
\begin{equation}\label{eq:cmop}
\begin{aligned}
\mathrm{Minimize} \qquad &\mathbf{F}(\mathbf{x}) =(f_{1}(\mathbf{x}), \cdots, f_{m}(\mathbf{x}))^{T}\\
\mathrm{subject\; to} \qquad  &\mathbf{x}\in \mathbb{R} \\
&{g}_{i}(\mathbf{x}) \leq 0,\; i=1,\cdots,p \\
&{h}_{j}(\mathbf{x}) =0,\; j=p+1,\cdots,q
\end{aligned},
\end{equation}
where $m$ denotes the number of objective functions; $\mathbf{x}=(x_1,\ldots,x_n)^T$ denotes the decision vector with $n$ dimensions (\ie~the number of decision variables); $\mathbf{x} \in \mathbb{R}$, and $\mathbb{R} \subseteq \mathbb{R}^n$ represents the search space. ${g}_{i}(\mathbf{x})$ and ${h}_{j}(\mathbf{x})$ are the $i$-th inequality and $j$-th equality constraints, and $q$ denotes the number of constraints.

In a CMOP, the degree of the $j$-th constraint violation (CV) of a solution $\mathbf{x}$ is
\begin{equation}\label{eq:cvj}
\varphi_{j}(\mathbf{x})=
\begin{cases}
    \mathop{\max}(0,g_{j}(\mathbf{x})), & j=1,\cdots,p \\
    \mathop{\max}(0,|h_{j}(\mathbf{x})|-\sigma), & j=p+1,\cdots,q
\end{cases},
\end{equation}
where $\sigma$ is a small enough positive value to relax the equality constraints into inequality ones. The overall CV of $\mathbf{x}$ is
{
\begin{equation}\label{eq:cv}
    \phi(\mathbf{x})=\sum^{q}_{j=1}\varphi_{j}(\mathbf{x}).
\end{equation}
}
$\mathbf{x}$ is feasible if {$\phi(\mathbf{x})=0$}; otherwise, it is infeasible.
}

\subsection{Adaptive Operator Selection in MOPs}\label{sec:related-work-os}

In recent years, adaptive operator selection is gradually attracting research attention in the design of multi-objective optimization evolutionary algorithms (MOEAs).

Wang~\etal~\cite{Wang2019Evolutionary} proposed a multi-operator ensemble method that uses multiple subpopulations for multiple operators and adjusts their sizes according to the effectiveness of each action to reward good operators and punish bad ones. Tian~\etal~\cite{Tian2022Deep} {adopted DRL to construct} an adaptive operator selection method for MOEA/D. The proposed DRL model regards the decision vectors as states and the operators as actions; then, the fitness improvement of the solution brought by the operator is taken as the reward. Alejandro~\etal~\cite{Alejandro2021Micro-Genetic} {suggested} a fuzzy selection of operators that chooses the most appropriate operators during evolution to promote both diversity and convergence of solutions. Dong~\etal~\cite{Dong2022Adaptive} {devised} a test-and-apply structure to adaptively select the operator for decomposition-based MOEAs. The proposed structure contains a test phase that uses all operators to generate offspring and test the survival rate as the credit of operators. Then, the apply phase uses only the best operator for the remaining evolution. Yuan~\etal~\cite{Yuan2015An} investigated the effect of different variation operators (\ie~GA and DE) in MOEA. McClymont~\etal~\cite{McClymont2011Markov} proposed a Markov chain hyper-heuristic that employs Reinforcement Learning and Markov chains to adaptively select heuristic methods, including different operators. Lin~\etal~\cite{Lin2022One-to-one} {devised} a one-to-one ensemble mechanism that measures the credits of operators in both the decision and objective spaces and designed an adaptive rule to guarantee that suitable operators can generate more solutions.

\subsection{DRL and its Applications in MOPs}\label{sec:related-work-drl}

As we know, Evolutionary Computation aims to find the optima from a static environment. However, Reinforcement Learning can learn an optimal coping strategy in a dynamic environment~\cite{Sutton1998Reinforcement}. The policy enables the agent to adaptively take actions that gain the largest cumulative reward at the current environmental state~\cite{Fayaz2022Machine}. Different from Reinforcement Learning, DRL uses a deep neural network to approximate the action-value function. More specifically, Q-Learning learns a policy through a discrete Q-Table that records the cumulative reward of each action at each state~\cite{Watkins1992Q-Learning}. In contrast, DQL trains a Q-Network that approximates the action-value function to estimate the expected reward of each action~\cite{Mnih2015Human-level}. The action-value function is also known as a probability model that estimates the probability of taking each action. Generally, DQL has two types of working principles.
\begin{figure}[!t]
	\centering
	\includegraphics[scale=0.5]{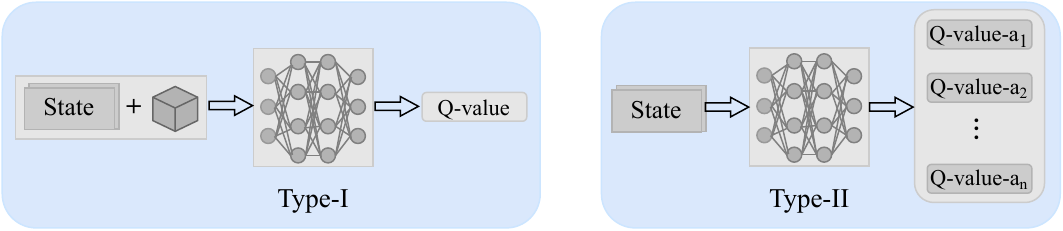}
	\caption{{An} illustration of two types of working principles of the DQL technique.}
	\label{fig:dql}
\end{figure}
We assume that the DQN is trained, where two types are depicted in the left half and right half of Fig~\ref{fig:dql}, respectively. For the Type-I DQL model, the input includes a state and an action (shown in a cube), and the output is the Q-value of this action under this state. For the Type-II DQL model, the input contains only the state, and the DQN can output the Q-values of all actions.

Without loss of generality, DQL trains a Q-Network through the loss function
\begin{equation}\label{eq:loss}
  \mathcal{L} = \frac{1}{|\mathcal{T}|}\sum_{\mathbf{t}\in \mathcal{T}}(Q(s_{t},a_{t})-q_{t})^2,
\end{equation}
where $\mathcal{T}$ is the training data randomly sampled from an \textbf{experience {replay}} (EP); the experience {replay} is a data set that records historical data for the training of the Network; $Q(s_{t},a_{t})$ is the output of the Network with the input $(s_t,a_t)$ and $a_T$, and $q_t$ is the Q-value of taking action $a_t$ at the state $s_t$, which is formulated as
\begin{equation}\label{eq:qt}
  q_t = r_t + \gamma \max_{a'\in \mathcal{A}} Q(s_{t+1},a'),
\end{equation}
where $Q(s_{t+1},a')$ is the maximum reward of taking the next action under the state where $a_t$ is performed. Therefore, $q_t$ {can} not only calculate the current reward of performing action $a$, but also estimate the maximum cumulative reward in the future.

Recently, DRL techniques are attracting more and more attention in the multi-objective optimization community. Researchers either employ DRL techniques to solve real-world MOPs or utilize them to enhance the performance of MOEAs.

\subsubsection{Applications in Real-world MOPs}\label{sec:related-work-drl-real}


Li~\etal~\cite{Li2022Many-Objective} {presented} a DRL-assisted multi-objective bacterial foraging optimization algorithm for a five-objective renewable energy accommodation problem. Luca~\etal~\cite{Caviglione2021Deep} {suggested} solving the multi-objective placement problem of virtual machines in cloud data centers by a DRL framework that selects the best placement heuristic for each virtual machine. Stefan~\etal~\cite{Schneider2021Self-Learning} {designed} a DRL-based DeepCoord approach that trains an agent to adaptively learn how to best coordinate services without prior and expert knowledge in dealing with a three-objective service coordination problem.

\subsubsection{Applications in MOEAs}\label{sec:related-work-drl-moea}

Tian~\etal~\cite{Tian2022Deep} {used the DRL technique to adaptively select} the evolutionary operators in dealing with MOPs. In their methods, the decision variable of a solution is regarded as the state, the operator is regarded as the action, and the improvement of fitness of the solution is the reward. Li~\etal~\cite{Li2021Deep} proposed modeling each subproblem decomposed by the reference vectors as a neural network and generating Pareto-optimal solutions directly through the trained network models. They modeled the multi-objective TSP (MOTSP) as a Pointer Network and solved it using DRL techniques. Zhang~\etal~\cite{Zhang2022Meta-Learning-Based} {developed} a concise meta-learning-based DRL technique that first trains a fine-tuned meta-model to derive submodels for the subproblems to shorten training time. Then, the authors further used their technique to solve MOTSP and multi-objective vehicle routing problems with time windows.

\subsection{Motivations}\label{sec:related-work-motivations}

In the Evolutionary Computation community, evolutionary operators (also known as variation operators) that generate new candidate solutions for selection are an important part~\cite{Tian2022Principled}. A fixed operator can not be suitable for all problems according to the No Free Lunch theorem. Therefore, adaptive operator selection is widely recognized as an interesting and key issue for Evolutionary Computation~\cite{Kerschke2019Automated}. Although adaptive operator selection has attracted some attention in the multi-objective optimization community, unfortunately, there are currently no research efforts dedicated to the design of CMOEAs. Given that a CMOP inherits the features and challenges of MOPs other than its constraints, it is valuable and promising to develop adaptive operator selection methods for solving CMOPs.

During the evolution of solving a CMOP, the whole environment is dynamic because the population is different and unknown in advance at each iteration. Therefore, adopting Reinforcement Learning techniques to solve the adaptive operator selection issue is intuitively effective. However, traditional Reinforcement Learning techniques such as Q-Learning may be less effective in dealing with such problems because the environment can include an infinite number of states, that is, the search space is continuous but Q-Table can only handle discrete state space. In contrast, DRL techniques that train a deep neural network as the policy are suitable for the continuous and infinite state space.

Another advantage of DRL techniques is that the output $q_t$ of Equation~\eqref{eq:qt} contains not only the current reward $r_t$ but also the maximum expected reward after taking this action. Consequently, $q_t$ can represent the maximum cumulative reward in the future due to the recursive nature of this formula~\cite{Tian2022Deep}. As a result, DRL techniques such as DQL are more suitable for the evolutionary process of CMOEAs since both historical and future performance should be considered in the heuristic algorithms~\cite{Gong2015Adaptive}. Thus, DQL is very promising for adaptive operator selection.

{Meanwhile, there are some challenging issues that must be resolved to successfully apply DRL to CMOPs. Since a CMOP contains constraints, it is necessary for a DRL model to consider constraint satisfaction and feasibility in the design of the state and reward. In addition, the effectiveness of an action (\ie~operator) on both handling constraints and optimizing objectives should be evaluated in training the DRL model.}

To adopt DQL in CMOEAs, we need to develop a model that determines the state, action, reward, and learning procedure. Our proposed DQL model and the DQL-assisted CMOEA framework are {elaborated on} in the next section.

\section{Proposed Methods}\label{sec:proposed}

\subsection{Proposed DRL Model}\label{sec:proposed-model}

In this work, the evolutionary operators are regarded as actions, thus the actions include
\begin{equation}\label{eq:action}
  {\mathcal{A} = \left\{ op_1, op_2, \cdots, op_i, \cdots, op_k \right\},}
\end{equation}
where $op_i$ is the $i$th evolutionary operator such as GA, DE, PSO, and CSO, and {$k$} is the number of candidate operators. It should be noted that in our proposed model, any number of operators can be used as candidates.

Then, we define the state of a population by considering its performance in terms of convergence, diversity, and feasibility. The average sum of objective functions of solutions in the population is used to evaluate the convergence of the population in the objective space,~\ie~the approximation to the CPF. Specifically, it is formulated as
\begin{equation}\label{eq:c}
  con = \frac{\sum_{\mathbf{x}\in \mathcal{P}} \sum_{j=1}^{m}f_{j}(\mathbf{x})}{N},
\end{equation}
where $f_{j}(\mathbf{x})$ is the $j$-th objective function value. When the population approximates the CPF, $con$ will become smaller. The objective functions are not normalized due to the following underlying considerations. After normalization, all objectives will belong to $[0,1]$. Although the influence of different scales can be eliminated, the $con$ fails to represent the real distribution of the population.

The average CV value of solutions in the population is used to estimate feasibility,~\ie~the distribution of the population in the feasible/infeasible regions. Specifically, it is formulated as
\begin{equation}\label{eq:f}
  fea = \frac{\sum_{\mathbf{x}\in \mathcal{P}} \phi(\mathbf{x})}{N},
\end{equation}
where $\phi(\mathbf{x})$ is the CV of $\mathbf{x}$. If all solutions of the population are in feasible regions, $fea$ is zero. Otherwise, it will be a large value if the population is located outside the feasible regions.

The sum of scales, in all objective dimensions that the population covers, is used to estimate the diversity of the population. Specifically, it is formulated as
\begin{equation}\label{eq:d}
  div = \frac{1}{\sum_{j=1}^{m} (f_{j}^{max} - f_{j}^{min})},
\end{equation}
where $f_{j}^{max}$ and $f_{j}^{min}$ represent the maximum and minimum objective function value on the $j$th objective function. If the population is stuck in a local feasible region, $div$ is very large because $f_{j}^{max}$ is close to $f_{j}^{min}$. On the contrary, if the population is well-distributed among every objective, $div$ will be small because $f_{j}^{max}$ is larger than $f_{j}^{min}$.

Then, we use three components to form a state. The state set is
\begin{equation}\label{eq:state}
  \mathcal{S} = \left\{ s| s=(con, fea, div) \right\}.
\end{equation}
It uses $con$, $fea$, and $div$ to reflect the current state of the population. In this case, whether an operator should be selected depends on a comprehensive evaluation of the current population state and its effectiveness in enhancing these performances.

The reward is calculated by
\begin{equation}\label{eq:reward}
  r = (con + fea + div) - (con' + fea' + div'),
\end{equation}
where $con'$, $fea'$, and $div'$ are the new state of the population for the next generation. The reward we define in this work represents the improvement of the population state, including the performance in terms of convergence, diversity, and feasibility. Therefore, the effectiveness of a selected operator can be comprehensively evaluated. {It is necessary to note that these three terms are normalized when they are input to the network as training/predicting data.}

A record is formed as
\begin{equation}\label{eq:record}
  \mathbf{t} = (s, op, r, s') = (con, fea, div, op, r, con', fea', div'),
\end{equation}
and the EP is formed as
\begin{equation}\label{eq:state}
  \mathcal{EP} = \left\{ \mathbf{t}_1, \mathbf{t}_2, \cdots, \mathbf{t}_{ms_{ep}} \right\}.
\end{equation}
Such a data structure enables the EP to record the state, the selected operator, the reward, and the new state at each iteration. Thus, when training the network, the training data include all necessary items to approximate the action-value function in handling the operator selection issue.


\begin{figure}[!t]
	\centering
	\includegraphics[scale=0.37]{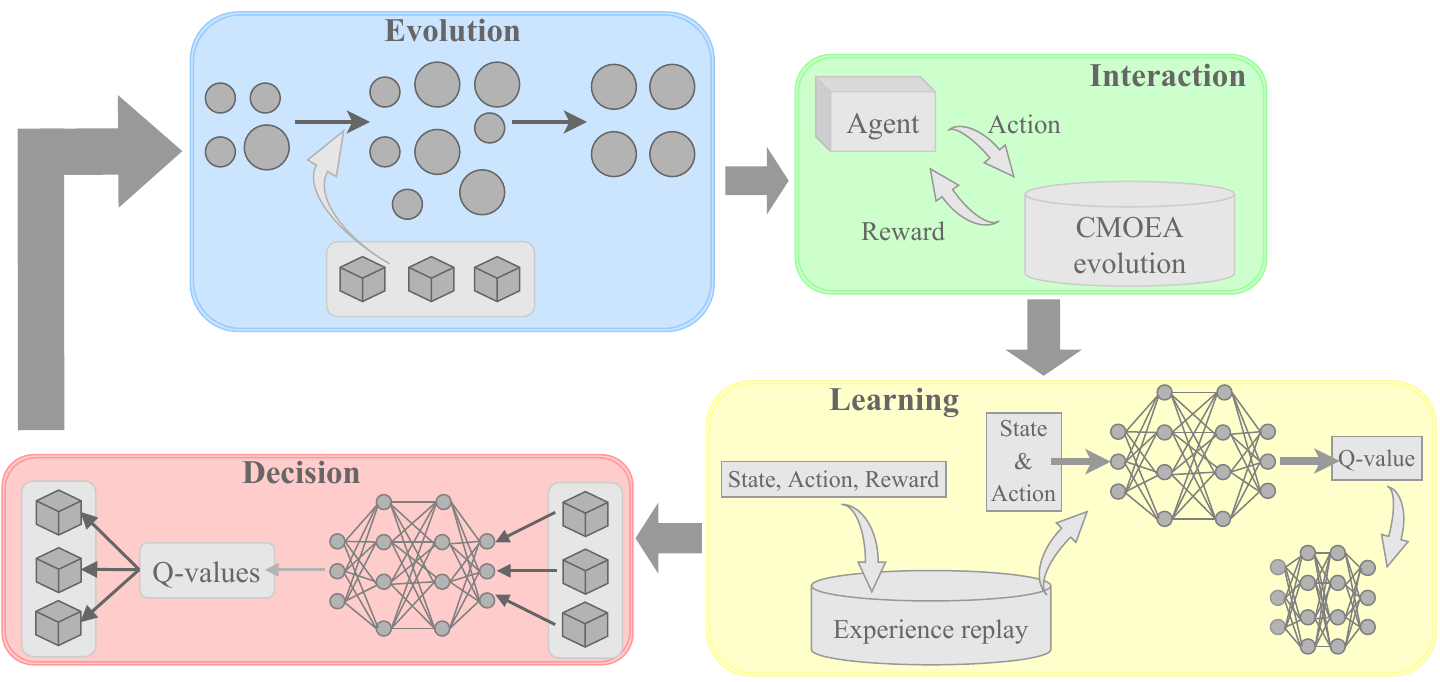}
	\caption{The illustration of the proposed DQL model.}
	\label{fig:model}
\end{figure}

Based on the above definitions and proposals, we develop a DQL model for operator selection for CMOEAs. The proposed model is shown in Fig.~\ref{fig:model}. This model contains four procedures, they are:
\begin{itemize}
 \item \emph{Evolution procedure:} In the evolution procedure, the CMOEA generates offspring by the selected operator (determined by the DQN) and determines solutions that can survive to the next generation.
 \item \emph{Interaction procedure:} In the interaction procedure, the agent and the environment exchange information and interact with each other. The agent gives the environment an action (\ie~operator) and the environment generates feedback (\ie~reward) from the evolution of the CMOEA for the agent.
 \item \emph{Learning procedure:} After the agent receives the feedback, it learns and improves the policy in the learning procedure. First, the record of the current iteration is added to the EP. Second, the DQN is trained based on the training data sampled from the EP. The input of the DQN is $State \& Action$, which is to say, the population state and the adopted operator at one iteration. The output of the DQN is the Q-value of adopting that operator at that population state.
 \item \emph{Decision procedure:} After the DQN is trained, the agent can use it to decide which action to take in {the face} of a population state. The decision procedure first estimates all Q-values of all actions. Then the action with the largest Q-value is selected.
\end{itemize}
The above four procedures execute in order at each iteration to achieve the DQL-assisted operator selection for CMOEA.

\begin{figure*}[!b]
	\centering
	\includegraphics[scale=0.7]{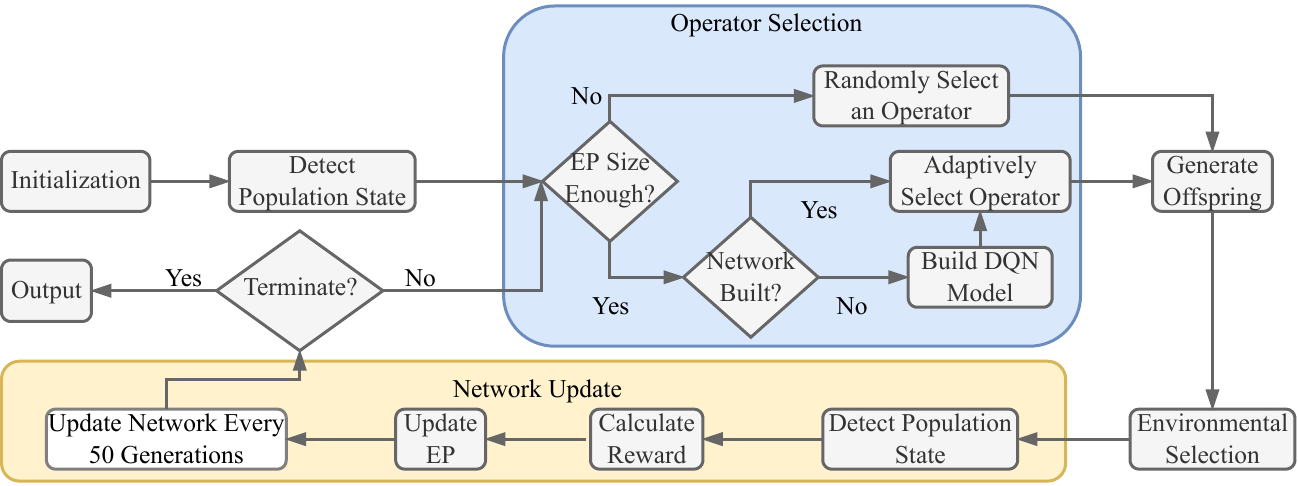}
	\caption{The flowchart of the proposed DQL-assisted CMOEA framework.}
	\label{fig:framework}
\end{figure*}

\subsection{Proposed DQL-assisted Framework}\label{sec:proposed-framework}

Based on the DQL model, we develop a DQL-assisted adaptive operator selection framework for CMOEAs. We present the flowchart of the proposed DQL-assisted CMOEA framework in Fig.~\ref{fig:framework}. The {main steps are as those in a CMOEA (initialization, mating, and selection)}, except that the DQL-assisted framework contains two additional parts.

The blue part contains the \textbf{Operator Selection} process. In this process, the size of the EP needs to meet the requirement before the DQN is trained because we need enough data to train the Network. If the size of the EP is not enough, we randomly select an operator at the iteration to give equal weight to each operator so that the exploration can be guaranteed. Otherwise, if the size of the EP meets the requirement and the DQN is not built, the DQN is first built. If the DQN has been trained, it is directly used to adaptively select an operator for the following steps.

After we get an adaptively selected operator, the offspring are generated based on the mating selection of the CMOEA, and the environmental selection of the CMOEA is conducted to determine the population for the next generation.

Then, the \textbf{Network Update} process (in yellow) is conducted. It mainly contains two specific parts. At each iteration, the population state is detected and the record is calculated. Afterward, the EP is updated by adding the record of this iteration. After every 50 iterations, the QDN is updated. The above steps continue until the termination condition is met.

\renewcommand{\algorithmicensure}{\textbf{Output:}}
\removelatexerror
\begin{algorithm}[!t]\small
\caption{The DQL-assisted Framework for CMOEAs}
\label{alg:framework-cmoos}
	\begin{algorithmic}[1]
	\REQUIRE $N$ (population size), $G_{max}$ (termination condition), $i$ (number of operators), $ms_{ep}$ (maximum size of EP), $rs_{ep}$ (required size of EP)
	\ENSURE  $\mathcal{P}$ (the output solution set of the CMOEA)
	\STATE Initialization of the CMOEA;
	\STATE Determine the state of the population;
	\STATE $\mathcal{EP} \leftarrow$ Initialize the EP;
	\STATE $g\leftarrow 0$;
	\WHILE {$g < G_{max}$}
	  \IF {$|\mathcal{EP}| < rs_{ep}$}
	    \STATE $i \leftarrow$ Randomly select an operator;
	    \STATE $\mathcal{O} \leftarrow$ Generate offspring set by the CMOEA using the $i$th operator;
	    \STATE $\mathcal{P} \leftarrow$ Select the population for the next generation by the CMOEA;
	    \STATE $\mathbf{t} \leftarrow$ Determine the reward and new state, and form a new record;
	    \STATE $\mathcal{EP} \leftarrow$ Update the EP with $\mathbf{t}$;
	  \ELSE
	    \IF {Network is not built}
	      \STATE $Q \leftarrow$ Build the DQN using $\mathcal{EP}$ by Algorithm~\ref{alg:train-network};
	    \ELSE
	      \STATE $i \leftarrow$ Adaptively select an operator according to the state of the population and $Q$ using Algorithm~\ref{alg:select-operator};
	      \STATE $\mathcal{O} \leftarrow$ Generate offspring set by the CMOEA using the $i$th operator;
	      \STATE $\mathcal{P} \leftarrow$ Select the population for the next generation by the CMOEA;
	      \STATE $\mathbf{t} \leftarrow$ Determine the reward and new state, and form a new record;
	      \STATE $\mathcal{EP} \leftarrow$ Update the EP with $\mathbf{t}$;
	    \ENDIF
	  \ENDIF
	  \STATE $g\leftarrow g+1$;
	  \IF {$g\% 50 = 0$}
	    \STATE $Q \leftarrow$ Update the DQN using $\mathcal{EP}$ by Algorithm~\ref{alg:train-network};
	  \ENDIF
	\ENDWHILE
	\RETURN $\mathcal{P}$
	\end{algorithmic}
\end{algorithm}

The Pseudocode of the proposed DQL-assisted framework for CMOEAs is presented in Algorithm~\ref{alg:framework-cmoos}. The inputs include the population size, the termination condition (maximum function evaluations), the maximum size of EP, and the required size of EP for training the DQN. The output is the same as the output solution set of the selected CMOEA. The main steps are consistent with the process in the flowchart. First, the initialization of the CMOEA is conducted to generate the initial population(s) for evolution (line 1). Then, the state of the initial population is determined (line 2) to prepare the first record. The EP set $\mathcal{EP}$ is initialized as empty then (line 3) and the index of the current generation $g$ is initialized as zero (line 4). In the main loop, the following steps are performed:
\begin{itemize}
 \item If the $\mathcal{EP}$ has not meet the required size $rs_{ep}$, lines 7-11 are performed. First, an operator (denoted as $i$th operator) is randomly selected (line 7). Then, the offspring set $\mathcal{O}$ is generated based on the operator and the mating selection of the CMOEA (line 8). Afterward, the environmental selection of the CMOEA is performed to select the population $\mathcal{P}$ for the next generation (line 9). According to this iteration, the reward and the new state are determined and used to form a new record $\mathbf{t}$ (line 10). Then $\mathcal{EP}$ is updated using the record $\mathbf{t}$ (line 11). It should be noted that $\mathcal{EP}$ is a queue that follows the \emph{first in, first out} rule.
 \item Otherwise if the $\mathcal{EP}$ has met the required size $rs_{ep}$, then lines 13-20 are performed. In the beginning, the DQN $Q$ is not built, so it is initialized using data from $\mathcal{EP}$ by Algorithm~\ref{alg:train-network} (line 14).
 \item In the later iterations where the DQN has been built, lines 16-20 are performed. First, an operator is adaptively selected according to the current state and the DQL-assisted method presented in Algorithm~\ref{alg:select-operator} (line 16). Then the offspring set $\mathcal{O}$ is generated by the CMOEA using the selected operator (line 17). Afterward, the population for the next generation is selected based on the environmental selection of the CMOEA (line 18). Finally, the new record is formed (line 19) and the $\mathcal{EP}$ is updated (line 20).
 \item {Once every} 50 iterations, the DQN is updated using $\mathcal{EP}$ by Algorithm~\ref{alg:select-operator} (line 25). This step guarantees that the DQN can approximate the recent environment.
\end{itemize}
Finally, the output solution set of the CMOEA is output as the final solution set (line 25).

\subsection{Train/Update the DQN}\label{sec:proposed-dqn}

\begin{table}[!t]
	\renewcommand{\arraystretch}{1.2}
	\centering
	\caption{Detailed parameter setting of the adopted DQN architecture.}
    \begin{tabular}{ll}
    \toprule
    Parameter & Value \\
    \midrule
    Number of hidden layers & 2 \\
    Number of neurons in hidden layers & 40 \\
    Number of nodes in input layer & 4 \\
    Number of nodes in output layer & 1 \\
    Batch size & $|\mathcal{T}|$ \\
    Maximum number of iterations & 80000 \\
    Decay of learning rate & 1.00E-04 \\
    Learning rate & 0.01 \\
    Bias from input to hidden layer & 0.1 \\
    Bias from hidden layer to output layer & 0 \\
    Activation function & ReLU \\
    \bottomrule
    \end{tabular}
  \label{tab:dqn-parameter}
\end{table}

In this work, we adopt a simple Back-propagation neural network as the DQN. The detailed parameter settings of the adopted DQN architecture are listed in Table~\ref{tab:dqn-parameter}. When the DQN is trained or updated, the procedure of Algorithm~\ref{alg:train-network} is conducted.

The inputs include the experience {replay} $\mathcal{EP}$ and the required size of training data $s_{tr}$. The output is a DQN $Q$. First, $s_{tr}$ records are sampled from $\mathcal{EP}$ and form a set $\mathcal{T}$ as the training data (line 1). Then, the former four items (\ie~the state $s_t$ and the action $a_t$) are used as the input of the DQN (lines 2-3), while the fifth item (\ie~the reward $r_t$) is used as the output of the DQN (line 4). The last three items are used to train DQN as the $s_{t+1}$ in Equation~\eqref{eq:qt} (line 5). An important step is that we need to normalize the inputs and output to eliminate the influence of different scales so that the DQN can be accurately estimated (line 6). Then, the normalized training data is used to train the DQN using Equation~\eqref{eq:loss} as the loss function (line 7).

\renewcommand{\algorithmicensure}{\textbf{Output:}}
\removelatexerror
\begin{algorithm}[!t]\small
\caption{Train/Update Network}
\label{alg:train-network}
	\begin{algorithmic}[1]
	\REQUIRE $\mathcal{EP}$ (experience {replay}), $s_{tr}$ (required size of training data)
	\ENSURE  $Q$ (DQN)
	\STATE $\mathcal{T} \leftarrow$ Randomly sample $s_{tr}$ records as the training data;
	\STATE $s_t \leftarrow \left\{ t_1, t_2, t_3 \right\}, \mathbf{t} \in \mathcal{T} $;
	\STATE $a_t \leftarrow t_4, \mathbf{t} \in \mathcal{T}$;
	\STATE $r_t \leftarrow t_5, \mathbf{t} \in \mathcal{T}$;
	\STATE $s_{t+1} \leftarrow \left\{ t_6, t_7, t_8 \right\}, \mathbf{t} \in \mathcal{T} $;
	\STATE Normalize all items of $\mathcal{T}$;
	\STATE $Q \leftarrow$ Train the DQN using Equation~\eqref{eq:loss} as loss function;
	\RETURN $Q$
	\end{algorithmic}
\end{algorithm}

\subsection{Proposed OS Method}\label{sec:proposed-os}

\renewcommand{\algorithmicensure}{\textbf{Output:}}
\removelatexerror
\begin{algorithm}[!t]\small
\caption{Select Operator}
\label{alg:select-operator}
	\begin{algorithmic}[1]
	\REQUIRE $\varepsilon$ (possibility of greedy), $s$ (current population state), $\mathcal{A}$ (operator set), $Q$ (DQN)
	\ENSURE  $a$ (the selected operator)
	\STATE $k \leftarrow$ Generate a random number in $[0,1]$;
	\IF {$k \le \varepsilon$}
	  \STATE $s \leftarrow$ Normalize all items of the state data $s$;
	  \STATE $i = \mathop{argmax}_{a\in \mathcal{A}} Q(s,a)$;
     \ELSE
       \STATE $i \leftarrow$ Generate a random number in $\left\{ 1,2,\cdots,{k} \right\}$;
     \ENDIF
	\RETURN the $i$th operator $a$ in $\mathcal{A}$
	\end{algorithmic}
\end{algorithm}

The OS method in this work is similar to the common method of selecting the action in Reinforcement Learning. The detailed Pseudocode is presented in Algorithm~\ref{alg:select-operator}. The inputs include the possibility of a greedy $\varepsilon$ to control the possibility of selecting the operator by DQN or in a random method. They also include the current population state $s$, the operator set $\mathcal{A}$, and the DQN $Q$. The output is the selected operator $a$. First, a random number $k$ between $[0,1]$ is generated (line 1). If $k$ is less than the possibility of greedy, the DQN is used to select an operator (lines 3-4). When DQN is used, all operators are tested through the DQN, and the one with the maximum reward is adopted. If $k$ is greater than $\varepsilon$, the index of the selected operator is randomly generated (line 6) to guarantee exploration. In this work, we use GA and DE operators (\ie~their crossover and mutation strategies) as the candidate actions to instantiate the framework. The reasons for selecting these two operators are two-fold:
\begin{itemize}
 \item First, GA and DE are the two most commonly used operators in existing CMOEAs;
 \item Second, GA can handle {multimodality} well and has a strong capability for convergence, while DE can handle linkages well and is good at exploring the decision {space~\cite{Tian2022Principled}}.
\end{itemize}

It is worth noting that before the $\mathcal{EP}$ meets the required size (\ie~line 6 in Algorithm~\ref{alg:framework-cmoos}), the operator is randomly selected to enhance exploration. Also, {parameter} $\varepsilon$ enables some random selection to enhance the exploration.

\subsection{Computational Complexity}\label{sec:proposed-complexity}

According to Algorithms~\ref{alg:framework-cmoos} to Algorithm~\ref{alg:select-operator}, the computational complexity of our methods is determined by three components,~\ie~the selected CMOEA, the training of DQN, and the calculations of population state. The complexity of training the DQN (a Backpropagation neural network) is $O(4u^2)$ if we use $u$ to represent the number of neurons. The time consumption of calculating the state is $O(mnN)$. The time complexity of a CMOEA is usually $O(mN^2)$ to $O(N^3)$ (\eg~CCMO~\cite{Tian2020Coevolutionary}). Therefore, the overall computational complexity is determined by the selected CMOEA.

\subsection{Remarks}\label{sec:proposed-remark}

The differences between our proposed operator selection method and existing {adaptive operator selection} methods are two-fold:
\begin{itemize}
 \item {Our method can apply to CMOPs because the design of the DQL model and the algorithmic framework both consider constraints and the feasibility of solutions, but existing methods are not applicable to CMOPs.}
 \item Our method is {based on DQL}, which can evaluate the improvement of the population brought by the selected operator in the {future, while} existing methods can only evaluate the improvement according to historical evolution.
\end{itemize}

Although Tian~\etal~\cite{Tian2022Deep} also proposed a DQL-assisted operator selection method, our methods are different from theirs in three aspects:
\begin{itemize}
 \item First, {this work handles CMOPs, while} the topic of~\cite{Tian2022Deep} is unconstrained MOPs. Handling CMOPs is more difficult than MOPs due to the existence of constraints.
 \item Second, our proposed DQL model is different from the model in~\cite{Tian2022Deep}. In~\cite{Tian2022Deep}, the decision variable is regarded as the {state, and} the improvement of a single solution (offspring \emph{vs} parent solution) is regarded as the reward.
 \item Third, the method in~\cite{Tian2022Deep} is a specific algorithm for MOPs, while our proposed framework can be embedded into any {CMOEAs}.
\end{itemize}

{
The differences between our methodology and existing applications of DRL in solving MOPs are as follows:
\begin{itemize}
 \item First, this work proposes a universal adaptive operator selection framework that can be used in any algorithm for CMOPs. By contrast, most existing applications focus on using DRL to solve a specific real-world application problem.
 \item Second, most existing applications of DRL in solving MOPs can only solve discrete MOPs, while our method can solve any form of CMOP as long as the embedded operators and the adopted CMOEA can handle discrete or continuous decision variables.
\end{itemize}
}

\section{Experimental Studies}\label{sec:experiments}

This section presents the experimental studies. Section~\ref{sec:experiments-settings} presents the detailed experimental settings. Section~\ref{sec:experiments-os} compares the DQL-assisted operator selection with the original CMOEA and random operator selection. Section~\ref{sec:experiments-comparison} gives the comparison studies between our method and state-of-the-art CMOEAs. {Section~\ref{sec:experiments-parameter} shows the parameter analyses of the DRL-assisted operator selection technique and the adopted neural network. Section~\ref{sec:experiments-ablation} studies the effectiveness of using two performance indicators, instead of the proposed simplistic assessments in Equations~\eqref{eq:c} and~\eqref{eq:d}, to assess the population state.}

\subsection{Experimental Settings}\label{sec:experiments-settings}

\subsubsection{Benchmark Problems}\label{sec:experiments-settings-benchmark}

In the experimental studies, we selected four {challenging CMOP benchmark} test suites to test our methods. The benchmarks include CF~\cite{Zhang2008Multiobjective}, DAS-CMOP~\cite{Fan2020Difficulty}, DOC~\cite{Liu2019Handling}, and LIR-CMOP~\cite{Fan2019An}. The main difficulties and challenges of these benchmarks are summarized in Table~\ref{tab:benchmarks} in the Supplementary file.

\subsubsection{Algorithms in Comparison}\label{sec:experiments-settings-algorithms}

In this work, we embed our proposed DQL-assisted operator selection framework into four existing CMOEAs; they are CCMO~\cite{Tian2020Coevolutionary}, MOEA/D-DAE~\cite{Zhu2020Constrained}, EMCMO~\cite{Qiao2022An}, and PPS~\cite{Fan2019Push}. In the comparison studies, we selected nine state-of-the-art CMOEAs for comparison, including methods using different operators. They are c-DPEA~\cite{Ming2021Dual-Population}, ToP~\cite{Liu2019Handling}, CMOEA-MS~\cite{Tian2021Balancing}, BiCo~\cite{Liu2021Handling}, MFOSPEA2~\cite{Jiao2022Multiform}, ShiP-A~\cite{Li2014Shift-Based}, DSPCMDE~\cite{Yu2022Dynamic}, NSGA-II-ToR~\cite{Ma2021New}, and CCEA~\cite{Yuan2022constrained}.

\subsubsection{Parameter Settings and Genetic Operators}\label{sec:experiments-settings-parameter}

For algorithms that use GA as the operator, the simulated binary crossover (SBX) and polynomial mutation (PM)~\cite{Agrawal2000Simulated} were used with the following parameter settings:
\begin{itemize}
  \item Crossover probability was $p_{c} = 1$; distribution index was $\eta_{c} = 20$;
  \item Mutation probability was $p_{m} = 1/n$; distribution index was $\eta_{m} = 20$.
\end{itemize}

For algorithms that use DE as the operator, the parameters $CR$ and $F$ in the DE operator were set to 1 and 0.5, respectively.

The evolutionary settings and parameters of our methods are presented in Table~\ref{tab:parameter} in the Supplementary file. The explanations for the settings are also presented in Section~\ref{supp-sec:parameter} in the Supplementary file. All other parameter settings of the comparison methods were the same as in their original literature (\ie~the default settings in PlatEMO).

\subsubsection{Performance Indicators}\label{sec:experiments-settings-indicators}

Inverted generational distance based on modified distance calculation (IGD+)~\cite{Ishibuchi2019Comparison} and hypervolume (HV)~\cite{Ishibuchi2019Comparison} were adopted as indicators to evaluate the performance of different algorithms. We used two Pareto-compliant indicators to achieve a sound and fair comparison~\cite{Ishibuchi2022Difficulties}. Detailed information on these three indicators could be found in the Supplementary file. The value $NaN$ means an algorithm cannot find a feasible solution or the final solution set is far from the CPF.

\begin{table}[!t]
	\renewcommand{\arraystretch}{1.1}
	\centering
	\caption{Average rankings and the p-values by the Friedman test on the effectiveness of DQL-assisted OS.}
	\resizebox{\linewidth}{!}
	{
	    \begin{tabular}{lllll}
	    \toprule
          & HV ranking & p-value & IGD+ ranking & p-value \\
          \midrule
    CCMO  & 2.5476 & \textbf{0.000001} & 2.5714 & \textbf{0.000001} \\
    RandOS-CCMO & 1.9762 & \textbf{0.021947} & 1.9524 & \textbf{0.029096} \\
    DRLOS-CCMO & \hl{1.4762} &       & \hl{1.4762} &  \\
    \midrule
          & HV ranking & p-value & IGD+ ranking & p-value \\
          \midrule
    EMCMO & 2.5000   & \textbf{0.000000}     & 2.5238 & \textbf{0.000000} \\
    RandOS-EMCMO & 2.1071 & \textbf{0.001063} & 2.1429 & \textbf{0.000208} \\
    DRLOS-EMCMO & \hl{1.3929} &       & \hl{1.3333} &  \\
    \midrule
          & HV ranking & p-value & IGD+ ranking & p-value \\
          \midrule
    MOEA/D-DAE & 2.3214 & \textbf{0.001873} & 2.2857 & \textbf{0.003220} \\
    RandOS-MOEA/D-DAE & 2.0357 & 0.071814 & 2.0714 & \textbf{0.049535} \\
    DRLOS-MOEA/D-DAE & \hl{1.6429} &       & \hl{1.6429} &  \\
    \midrule
          & HV ranking & p-value & IGD+ ranking & p-value \\
          \midrule
    PPS   & 1.9881 & \textbf{0.010346} & 2.0476 & \textbf{0.012090} \\
    RandOS-PPS & 2.2857 & 0.230062 & 2.2500  & 0.113631 \\
    DRLOS-PPS & \hl{1.7262} &       & \hl{1.7024} &  \\
    \bottomrule
    \end{tabular}
	}
\label{tab:friedman}
\end{table}

\begin{table*}[!t]
	\renewcommand{\arraystretch}{1.1}
	\centering
	\caption{Statistical results of IGD+ obtained by DRLOS-EMCMO and other methods on DOC benchmark problems. The best result in each row is highlighted.}
	\resizebox{\textwidth}{!}
	{
\begin{tabular}{ccccccccccc}
\toprule
Problem&cDPEA&ToP&CMOEA\_MS&BiCo&MFOSPEA2&ShiP\_A&DSPCMDE&NSGAIIToR&CCEA&DRLOS-EMCMO\\
\midrule
\multirow{1}{*}{DOC1}&5.5922e-1 (5.87e-1) $-$&3.3066e-3 (1.48e-4) $-$&3.9980e+0 (3.22e+0) $-$&2.5299e-2 (4.21e-2) $-$&1.2071e-1 (2.31e-1) $-$&2.5551e+0 (2.06e+0) $-$&3.9738e+2 (4.57e+2) $-$&6.3475e+1 (2.25e+1) $-$&1.6477e+0 (1.59e+0) $-$&\hl{2.6377e-3 (1.82e-4)}\\
\hline
\multirow{1}{*}{DOC2}&NaN (NaN)&NaN (NaN)&NaN (NaN)&NaN (NaN)&NaN (NaN)&NaN (NaN)&2.4608e-1 (7.76e-2) $-$&NaN (NaN)&NaN (NaN)&\hl{4.3113e-2 (7.17e-2)}\\
\hline
\multirow{1}{*}{DOC3}&7.4298e+2 (2.24e+2) $\approx$&1.8914e+2 (1.74e+2) $+$&5.9087e+2 (2.60e+2) $\approx$&4.9838e+2 (2.67e+2) $\approx$&7.3066e+2 (2.74e+2) $\approx$&NaN (NaN)&\hl{1.1650e+2 (1.66e+2) $+$}&NaN (NaN)&5.8119e+2 (1.29e+2) $\approx$&6.0270e+2 (4.85e+2)\\
\hline
\multirow{1}{*}{DOC4}&6.0555e-1 (3.81e-1) $-$&1.0471e-1 (7.36e-2) $\approx$&7.7869e-1 (5.33e-1) $-$&2.5349e-1 (1.82e-1) $-$&3.9220e-1 (3.13e-1) $-$&8.6957e-1 (8.10e-1) $-$&4.4231e-2 (8.64e-2) $-$&1.6298e+1 (6.10e+0) $-$&8.7862e-1 (5.61e-1) $-$&\hl{4.1216e-2 (4.26e-2)}\\
\hline
\multirow{1}{*}{DOC5}&NaN (NaN)&3.8442e+1 (6.64e+1) $-$&9.5192e+1 (2.16e+1) $-$&NaN (NaN)&NaN (NaN)&NaN (NaN)&NaN (NaN)&NaN (NaN)&NaN (NaN)&\hl{2.1808e+1 (5.00e+1)}\\
\hline
\multirow{1}{*}{DOC6}&3.5707e+0 (2.89e+0) $-$&5.9406e+0 (2.48e+0) $-$&2.4328e+0 (2.55e+0) $-$&9.1093e-1 (8.16e-1) $-$&9.6498e-1 (1.13e+0) $-$&1.8291e+0 (1.77e+0) $-$&\hl{2.7957e-3 (1.78e-4) $\approx$}&2.3044e+1 (6.45e+0) $-$&2.6678e+0 (2.51e+0) $-$&9.4061e-3 (3.24e-2)\\
\hline
\multirow{1}{*}{DOC7}&7.3217e+0 (2.11e+0) $-$&1.3820e+0 (7.42e-1) $-$&3.8020e+0 (1.32e+0) $-$&5.2266e+0 (2.01e+0) $-$&5.3131e+0 (2.15e+0) $-$&NaN (NaN)&2.7176e-1 (8.13e-1) $-$&NaN (NaN)&5.5487e+0 (2.08e+0) $-$&\hl{1.4072e-1 (2.67e-1)}\\
\hline
\multirow{1}{*}{DOC8}&7.1094e+1 (4.52e+1) $-$&5.9654e+1 (2.73e+1) $-$&1.6705e+2 (7.52e+1) $-$&6.4607e+1 (5.86e+1) $-$&7.8331e+1 (6.07e+1) $-$&8.9302e+1 (5.55e+1) $-$&\hl{2.1322e-1 (5.80e-2) $+$}&3.7281e+2 (8.39e+1) $-$&8.9903e+1 (5.45e+1) $-$&2.7264e-1 (8.88e-2)\\
\hline
\multirow{1}{*}{DOC9}&1.5307e-1 (1.16e-1) $\approx$&2.2487e-1 (7.38e-2) $-$&7.9311e-2 (1.06e-1) $-$&1.4714e-1 (1.01e-1) $-$&1.1725e-1 (9.59e-2) $\approx$&1.3416e-1 (1.12e-1) $-$&8.8190e-2 (1.05e-2) $-$&6.3222e-1 (1.41e-1) $-$&1.5036e-1 (1.24e-1) $\approx$&\hl{5.0283e-2 (9.22e-3)}\\
\hline
\multicolumn{1}{c}{$+/-/\approx$}&0/5/2&1/6/1&0/7/1&0/6/1&0/5/2&0/5/0&2/5/1&0/5/0&0/5/2&\\
\bottomrule
\end{tabular}
}
  \label{tab:comparison-doc-igdp}
\end{table*}

\subsubsection{Statistical Analysis}\label{sec:experiments-settings-statistical}

Each algorithm is executed $30$ independent runs on each test instance. The mean and standard deviation values of IGD+ and HV were recorded. The Wilcoxon rank-sum test with a significance level of $0.05$ was used to perform the statistical analysis using the KEEL software~\cite{KEEL}. ``$+$'', ``$-$'', and ``$=$'' were used to show that the result of other algorithms was significantly better than, significantly worse than, and statistically similar to those obtained by our methods to the Wilcoxon test, respectively. All $NaN$ values of HV and IGD+ are replaced by zero and 100, respectively.

\begin{table*}[!t]
	\renewcommand{\arraystretch}{1.1}
	\centering
	\caption{Statistical results of HV obtained by DRLOS-EMCMO and other methods on LIR-CMOP benchmark problems. The best result in each row is highlighted.}
	\resizebox{\textwidth}{!}
	{
\begin{tabular}{ccccccccccc}
\toprule
Problem&cDPEA&ToP&CMOEA\_MS&BiCo&MFOSPEA2&ShiP\_A&DSPCMDE&NSGAIIToR&CCEA&DRLOS-EMCMO\\
\midrule
\multirow{1}{*}{LIRCMOP1}&1.6448e-1 (1.08e-2) $+$&1.0538e-1 (8.84e-3) $-$&1.0477e-1 (1.34e-2) $-$&1.3536e-1 (6.52e-3) $\approx$&1.3108e-1 (1.60e-2) $-$&\hl{2.2967e-1 (2.04e-3) $+$}&1.9200e-1 (2.37e-2) $+$&9.6272e-2 (5.01e-3) $-$&1.6415e-1 (1.47e-2) $+$&1.4210e-1 (1.90e-2)\\
\hline
\multirow{1}{*}{LIRCMOP2}&2.8144e-1 (1.58e-2) $+$&2.1555e-1 (1.42e-2) $-$&2.2253e-1 (2.32e-2) $-$&2.5522e-1 (1.01e-2) $\approx$&2.5272e-1 (1.53e-2) $\approx$&\hl{3.5009e-1 (2.21e-3) $+$}&3.2355e-1 (1.62e-2) $+$&2.0338e-1 (6.34e-3) $-$&2.9124e-1 (1.40e-2) $+$&2.6388e-1 (2.75e-2)\\
\hline
\multirow{1}{*}{LIRCMOP3}&1.4625e-1 (1.43e-2) $+$&9.2802e-2 (5.34e-3) $-$&9.9956e-2 (1.62e-2) $-$&1.2583e-1 (6.96e-3) $+$&1.1792e-1 (1.17e-2) $\approx$&\hl{1.9042e-1 (6.25e-3) $+$}&1.6138e-1 (2.11e-2) $+$&8.7510e-2 (4.22e-3) $-$&1.5215e-1 (1.21e-2) $+$&1.1822e-1 (1.61e-2)\\
\hline
\multirow{1}{*}{LIRCMOP4}&2.4589e-1 (1.43e-2) $+$&1.8034e-1 (9.33e-3) $-$&1.8730e-1 (1.51e-2) $-$&2.1935e-1 (1.32e-2) $\approx$&2.1831e-1 (1.16e-2) $\approx$&\hl{2.8717e-1 (1.10e-2) $+$}&2.6399e-1 (2.49e-2) $+$&1.7739e-1 (6.98e-3) $-$&2.5082e-1 (1.34e-2) $+$&2.1807e-1 (1.98e-2)\\
\hline
\multirow{1}{*}{LIRCMOP5}&1.5599e-1 (2.41e-2) $-$&0.0000e+0 (0.00e+0) $-$&1.0881e-1 (6.93e-2) $-$&0.0000e+0 (0.00e+0) $-$&1.4741e-1 (2.38e-2) $-$&1.4847e-1 (1.58e-2) $-$&2.6549e-1 (3.49e-2) $-$&0.0000e+0 (0.00e+0) $-$&8.9994e-2 (7.55e-2) $-$&\hl{2.8301e-1 (1.38e-2)}\\
\hline
\multirow{1}{*}{LIRCMOP6}&1.0447e-1 (1.26e-2) $-$&3.8214e-3 (1.45e-2) $-$&6.4431e-2 (5.03e-2) $-$&0.0000e+0 (0.00e+0) $-$&1.0871e-1 (1.33e-2) $-$&9.8962e-2 (9.44e-3) $-$&1.6168e-1 (4.32e-2) $-$&0.0000e+0 (0.00e+0) $-$&5.1631e-2 (4.62e-2) $-$&\hl{1.9379e-1 (1.07e-3)}\\
\hline
\multirow{1}{*}{LIRCMOP7}&2.5030e-1 (7.64e-3) $-$&1.4326e-2 (5.45e-2) $-$&2.4401e-1 (7.28e-3) $-$&1.6982e-1 (1.13e-1) $-$&2.5122e-1 (8.92e-3) $-$&2.4627e-1 (9.69e-3) $-$&2.8984e-1 (1.05e-2) $-$&0.0000e+0 (0.00e+0) $-$&2.4328e-1 (7.28e-3) $-$&\hl{2.9240e-1 (4.63e-3)}\\
\hline
\multirow{1}{*}{LIRCMOP8}&2.3549e-1 (9.02e-3) $-$&1.3858e-2 (5.27e-2) $-$&2.2602e-1 (9.02e-3) $-$&5.9252e-2 (1.00e-1) $-$&2.3840e-1 (1.01e-2) $-$&2.3424e-1 (1.20e-2) $-$&2.9288e-1 (4.64e-4) $-$&0.0000e+0 (0.00e+0) $-$&2.2442e-1 (4.88e-3) $-$&\hl{2.9365e-1 (9.30e-4)}\\
\hline
\multirow{1}{*}{LIRCMOP9}&3.7822e-1 (6.59e-2) $-$&2.9975e-1 (7.81e-2) $-$&2.3427e-1 (6.54e-2) $-$&1.2413e-1 (4.59e-2) $-$&3.2135e-1 (6.48e-2) $-$&3.4557e-1 (5.83e-2) $-$&3.4121e-1 (2.58e-2) $-$&3.6661e-2 (1.23e-2) $-$&2.2556e-1 (5.96e-2) $-$&\hl{4.4138e-1 (2.17e-2)}\\
\hline
\multirow{1}{*}{LIRCMOP10}&5.4129e-1 (5.30e-2) $-$&4.5018e-1 (1.02e-1) $-$&3.6909e-1 (1.67e-1) $-$&6.4856e-2 (3.07e-2) $-$&5.0681e-1 (9.87e-2) $-$&3.5948e-1 (1.05e-1) $-$&5.9874e-1 (2.98e-2) $-$&5.3776e-2 (2.94e-2) $-$&1.2244e-1 (9.89e-2) $-$&\hl{6.8631e-1 (1.23e-2)}\\
\hline
\multirow{1}{*}{LIRCMOP11}&6.3446e-1 (3.25e-2) $-$&3.8783e-1 (8.40e-2) $-$&3.9226e-1 (1.19e-1) $-$&2.2469e-1 (8.90e-2) $-$&6.2055e-1 (3.97e-2) $-$&4.5297e-1 (1.03e-1) $-$&6.2135e-1 (5.89e-2) $-$&6.0264e-2 (2.87e-2) $-$&3.3593e-1 (1.63e-1) $-$&\hl{6.8144e-1 (9.55e-3)}\\
\hline
\multirow{1}{*}{LIRCMOP12}&5.2337e-1 (4.27e-2) $-$&4.7168e-1 (4.51e-2) $-$&4.2431e-1 (7.04e-2) $-$&3.4807e-1 (1.07e-1) $-$&5.1668e-1 (3.92e-2) $-$&4.9940e-1 (3.07e-2) $-$&5.1211e-1 (5.02e-2) $-$&7.5184e-2 (2.28e-2) $-$&4.1243e-1 (5.49e-2) $-$&\hl{5.7470e-1 (1.69e-2)}\\
\hline
\multirow{1}{*}{LIRCMOP13}&5.5469e-1 (1.72e-3) $+$&2.3424e-3 (1.26e-2) $-$&5.4982e-1 (3.00e-2) $+$&1.1278e-4 (1.46e-4) $-$&1.1147e-4 (1.02e-4) $-$&5.3448e-1 (3.24e-3) $\approx$&5.0437e-1 (3.10e-3) $-$&0.0000e+0 (0.00e+0) $-$&\hl{5.5860e-1 (9.25e-4) $+$}&5.3441e-1 (3.20e-3)\\
\hline
\multirow{1}{*}{LIRCMOP14}&5.5445e-1 (1.37e-3) $+$&2.4647e-3 (1.23e-2) $-$&5.5565e-1 (1.14e-3) $+$&3.9193e-4 (3.00e-4) $-$&4.6467e-4 (2.92e-4) $-$&5.4340e-1 (2.85e-3) $-$&5.3045e-1 (3.44e-3) $-$&5.3460e-5 (1.79e-4) $-$&\hl{5.5810e-1 (7.14e-4) $+$}&5.4856e-1 (1.82e-3)\\
\hline
\multicolumn{1}{c}{$+/-/\approx$}&6/8/0&0/14/0&2/12/0&1/10/3&0/11/3&4/9/1&4/10/0&0/14/0&6/8/0&\\
\bottomrule
\end{tabular}
}
  \label{tab:comparison-lir-hv}
\end{table*}

\subsection{On the Effectiveness of DQL-assisted OS}\label{sec:experiments-os}

In the first part of the experiments, we compare the CMOEAs embedded using the DQL-assisted operator selection method with the original CMOEAs using a fixed operator and the CMOEAs using random operator selection to verify the effectiveness of the DQL-assisted operator selection method. The results in terms of HV and IGD+ on the four benchmarks are presented in Tables~\ref{tab:embed-cf-hv} to~\ref{tab:embed-lir-igdp} in the Supplementary file.

For the CFs, it can be found that DRLOS-CCMO and DRLOS-EMCMO significantly outperformed the corresponding methods using a fixed operator or random operator. {However, CCMO and EMCMO perform better or at least competitively on the three-objective CF8-10, revealing that for these large objective space instances, the operators of GA can enhance convergence.} As for MOEA/D-DAE and PPS, it can be determined that using random and adaptively selected operators outperformed using a fixed operator. For DAS-CMOPs, the superiority of CCMO and EMCMO on DAS-CMOP4-8 reveals that these instances prefer the GA operator. On the contrary, DAS-CMOP1-3 and DAS-CMOP9 prefer the DE operator. However, an adaptive operator selection method can better handle DAS-CMOP1-3 and DAS-CMOP9 than PPS using the DE operator, revealing that the DQL-assisted operator selection can learn to decide which operator to use according to the population state during evolution. For DOCs, the results are similar to those of CFs. DRLOS-CCMO and DRLOS-EMCMO performed significantly better than other methods, revealing that DOC prefers the DE operator and the adaptive selection method can accurately determine DE as the operator. {Since PPS uses DE as the operator, the performance is not significantly influenced by operator selection on DOCs. The results among RandOS and DRLOS also reveal that using DRL to adaptively select operators during evolution performs better than random selection.} For LIR-CMOPs, it is apparent that the proposed OS method can significantly improve the performances of all CMOEAs. {Nevertheless, the results show that LIR-CMOP13-14 prefer GA as operator.}

To better understand the performance, we conduct the Friedman test with Holm correction at a significance level of 0.05 on all results. The average rankings and p-values are summarized in Table~\ref{tab:friedman}. It can be found that the DQL-assisted operator selection method outperforms the fixed operator and random operator selection. For these four CMOEAs, using our proposed DQL-assisted operator selection method can improve their performance.

In summary, our proposed DQL-assisted OS method can improve the performance of these CMOEAs. The DQL-assisted operator selection method can better determine the operator based on the population state compared to using a fixed or randomly selected operator.

\begin{table}[!t]
	\renewcommand{\arraystretch}{1.1}
	\centering
	\caption{Average rankings and the p-values by the Friedman test on the comparison studies.}
	\resizebox{\linewidth}{!}
	{
    \begin{tabular}{lllll}
    \toprule
          & HV ranking & p-value & IGD+ ranking & p-value \\
          \midrule
    c-DPEA & 4.2738  & \textbf{0.027929}  & 4.2381  & \textbf{0.020103}  \\
    ToP   & 7.1548  & \textbf{0.000000}  & 7.2024  & \textbf{0.000000}  \\
    CMOEA-MS & 6.4286  & \textbf{0.000000}  & 6.4762  & \textbf{0.000000}  \\
    BiCo  & 6.6071  & \textbf{0.000000}  & 6.3690  & \textbf{0.000000}  \\
    MFOSPEA2 & 4.8929  & \textbf{0.001717}  & 5.0357  & \textbf{0.000413}  \\
    ShiP-A & 4.5833  & \textbf{0.007658}  & 4.4881  & \textbf{0.006876}  \\
    DSPCMDE & 3.8690  & \textbf{0.040718}  & 3.9981  & \textbf{0.034346}  \\
    NSGA-II-ToR & 9.3333  & \textbf{0.000000}  & 9.5238  & \textbf{0.000000}  \\
    CCEA  & 5.5357  & \textbf{0.000040}  & 5.4762  & \textbf{0.000027}  \\
    DRLOS-EMCMO & \hl{2.8214}  &       & \hl{2.7024}  &  \\
    \bottomrule
    \end{tabular}
	}
\label{tab:friedman-comparison}
\end{table}

\begin{figure*}[!t]
	\centering
	\subfigure[]{ \includegraphics[scale=0.55]{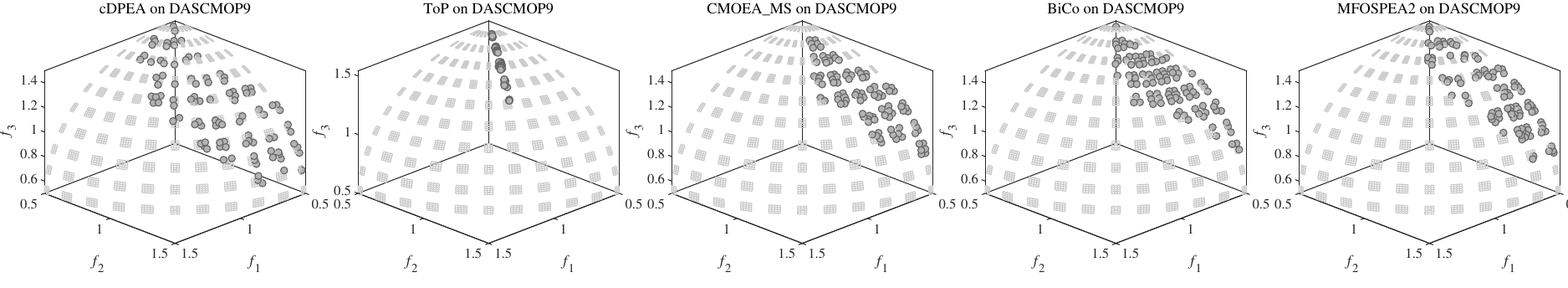}}
	\subfigure[]{ \includegraphics[scale=0.55]{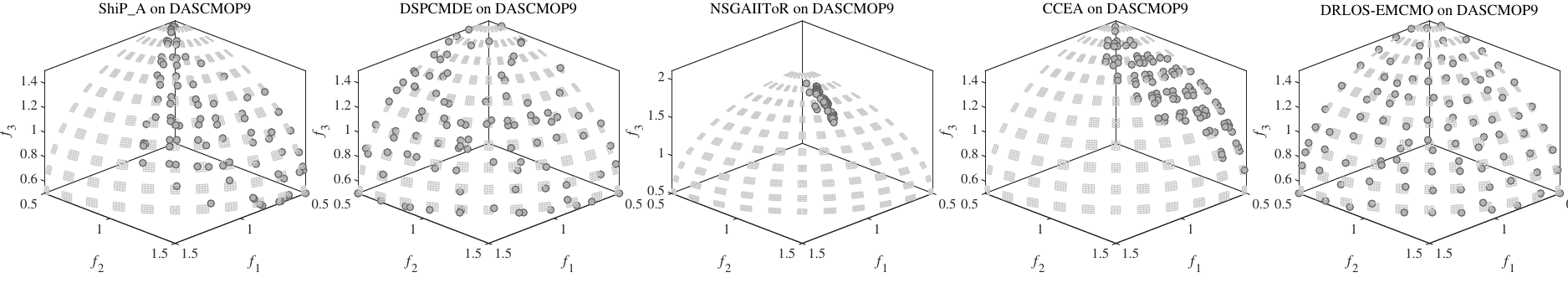}}
	\caption{The final solution sets obtained by DRLOS-EMCMO and other methods on DAS-CMOP9 with the median IGD+ value among $30$ runs.}
	\label{fig:cpf-cmparison-das9}
\end{figure*}

\subsection{Comparison Studies}\label{sec:experiments-comparison}

After the verification of the proposed DQL-assisted OS method, we further compare the DRLOS-EMCMO algorithm with nine state-of-the-art CMOEAs to study the algorithmic performance of these CMOEAs. The statistical results on four benchmarks are presented in Tables~\ref{tab:comparison-doc-igdp} to~\ref{tab:comparison-lir-hv} in this file and Tables~\ref{tab:comparison-cf-hv} to~\ref{tab:comparison-lir-igdp} in the Supplementary file.

In general, DRLOS-EMCMO outperforms other CMOEAs on these challenging CMOPs. But it performed worse in some CMOPs. {Among the selected CMOPs, some instances rely on a particular operator. For example, the results show that DAS-CMOP4-8 and LIR-CMOP13-14 need the GA operator because of their multimodal feature. Besides, some instances of the CF test suite need the DE operator due to the linkage between variables. However, since the computational resources are used for all operators in DRLOS-EMCMO in the learning stage, DRLOS-EMCMO has fewer resources for the specifically needed operator in dealing with these problems, and thus, it performs worse. Additionally, it can be found that LIR-CMOP1-4, the CMOPs with an extremely small feasible region, require some constraint relaxation techniques such as penalty function in c-DPEA~\cite{Ming2021Dual-Population} and ShiP-A~\cite{Ma2021Shift}, or $\varepsilon$-constrained in DSPCMDE~\cite{Yu2022Dynamic} and CCEA~\cite{Yuan2022constrained}. However, EMCMO does not contain a constraint relaxation technique, and thus, it performs worse on LIR-CMOP1-4. Except for these specific problems, DRLOS-EMCMO generally performs better than other CMOEAs.}

We depict the final solutions sets obtained by DRLOS-EMCMO and other CMOEAs on CF1, DAS-CMOP9, DOC2, and LIR-CMOP8 in Figs~\ref{fig:cpf-cmparison-cf1} to~\ref{fig:cpf-cmparison-lir8} in the Supplementary file and Fig~\ref{fig:cpf-cmparison-das9} in this file. The results clearly show that DRLOS-EMCMO can approximate the CPF and obtain an even distribution in every instance. {For DOC2, it can be found that only DRLOS-EMCMO can converge to two segments of the CPF, revealing that using GA performs worse than using an adaptively selected operator. As for LIR-CMOP8, it can be found DSPCMDE, adopting DE as the operator, also converges to the CPF. However, some dominant and extreme solutions remain, revealing that when GA can be adaptively selected, convergence can be further enhanced. Similarly for DAS-CMOP9, DSPCMDE finds fewer segments of the CPF compared to DRLOS-EMCMO, demonstrating that when GA can be used during evolution, the population can more effectively converge to the CPF and is less likely to get trapped in local optima.}

\begin{figure}[htb]
	\centering
	\subfigure[]{ \includegraphics[scale=0.57]{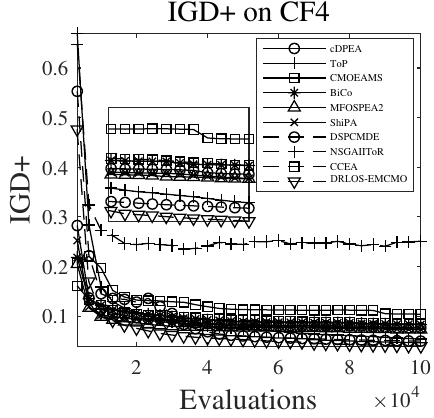}\label{fig:comparison-igdp-convergence-1}}
	\subfigure[]{ \includegraphics[scale=0.57]{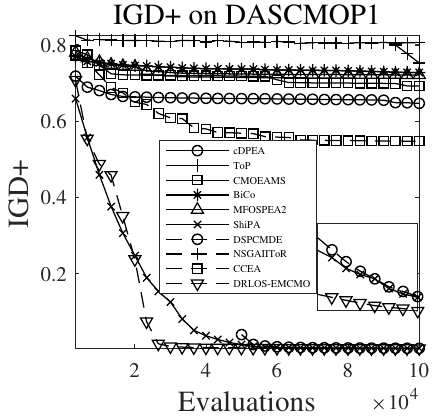}\label{fig:comparison-igdp-convergence-2}}
	\subfigure[]{ \includegraphics[scale=0.57]{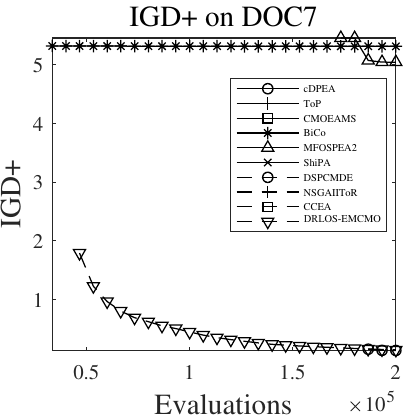}\label{fig:comparison-igdp-convergence-3}}
	\subfigure[]{ \includegraphics[scale=0.57]{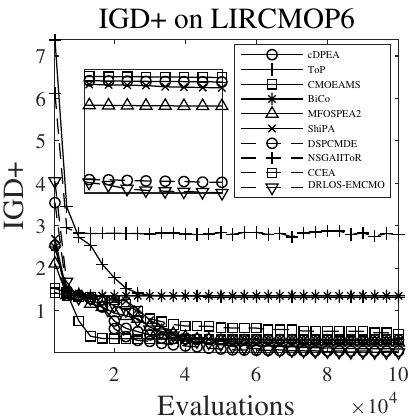}\label{fig:comparison-igdp-convergence-4}}
	\caption{The convergence profiles on IGD+ of DRLOS-EMCMO and other methods on CF4, DAS-CMOP1, DOC7, and LIR-CMOP6 with the median IGD+ values among $30$ runs.}
	\label{fig:comparison-igdp-convergence}
\end{figure}

Additionally, we depict the convergence profiles of DRLOS-EMCMO and other CMOEAs on CF4, DAS-CMOP1, DOC7, and LIR-CMOP6 in Fig~\ref{fig:comparison-igdp-convergence}. It can be determined that DRLOS-EMCMO can not only achieve faster convergence speed but also obtain a better final indicator value.

In summary, DRLOS-EMCMO outperforms these CMOEAs using a fixed operator, revealing that the DQL-assisted operator selection can achieve better versatility on different CMOPs.

{
\subsection{Parameter Analyses}\label{sec:experiments-parameter}

In the first part of parameter analyses, we change the required size of EP and the batch size to test its influence on training the DQL. Also, we change the greedy threshold to test the robustness of our method. The statistical results are presented in Tables~\ref{tab:parameter-ccmo-cf-hv} to~\ref{tab:parameter-pps-lir-igdp} in the Supplementary file. It can be found that these two parameters have little influence on the performance in terms of all benchmark problems, demonstrating that our method is not parameter sensitive. Therefore current parameter settings are applicable when applying the DQL-assisted operator selection method to a new CMOEA.

Then in the second part of parameter analyses, we conduct different settings of important parameters in the DQN, including the decay of learning rate, learning rate, number of iterations in training, and number of nodes in each hidden layer, to test the influence of these parameter settings. Detailed explanations of variants are reported in Table~\ref{tab:name} in the Supplementary file. In this part, only DRLOS-EMCMO is adopted to save space. Statistical results of HV and IGD+ are reported in Tables~\ref{tab:dqn-parameter-cf-hv} to~\ref{tab:dqn-parameter-lir-igdp} in the  Supplementary file. On all benchmark problems, different parameter settings have little or no significant influence on DRLOS-EMCMO, revealing that the DQN adopted in this work is not parameter sensitive.

\subsection{Ablation Studies on Assessing Population State Using Indicators}\label{sec:experiments-ablation}
In this part, we adopt two indicators, HV~\cite{Ishibuchi2019Comparison} and Spacing~\cite{Schott1995Fault}, to estimate the population state, respectively. These two indicators are selected because they do not need prior knowledge of the true PF. Compared to the items used in Equations~\eqref{eq:c} and~\eqref{eq:d}, HV and Spacing are two more sophisticated indicators that can provide a more concise estimation of the population state. The HV indicator can evaluate convergence and diversity, while the Spacing indicator can evaluate diversity.

The results are reported in Tables~\ref{tab:indicator-hv} and~\ref{tab:indicator-igdp} in the Supplementary file, according to which the following conclusions can be found.
\begin{itemize}
 \item For CCMO and EMCMO, using simplistic evaluations as in Equations~\eqref{eq:c} and~\eqref{eq:d} performs better when dealing with DSA-CMOP1-3 and DOCs, revealing that a sophisticated evaluation can easily lead to local optima because these instances need not only operators of GA but also other operators. In contrast, using HV and Spacing indicators performs better on DAS-CMOP4-9 which prefers GA. The indicators provide a more concise estimation of the dynamics of the population, allowing the algorithm to determine how to select operators of GA.
 \item For MOEA/D-DAE, the choice of simplistic or sophisticated evaluations has no significant influence on the performance.
 \item For PPS, using sophisticated evaluations performs significantly worse on instances with no preference for operators of GA and performs better on some instances with a preference for operators of GA, which further demonstrates that a simplistic evaluation has better versatility.
\end{itemize}
}

\section{Conclusions and Future Work}\label{sec:conclusions}

In this article, we propose a DQL-assisted online operator selection method for CMOPs, filling the research gap in operator selection in CMOPs and introducing DRL techniques to CMOPs. We develop a DQL model that uses population convergence, diversity, and feasibility as the state, the operators as actions, and the improvement of the population state as the reward. Based on this DQL model, we develop a versatile and easy-to-use DQL-assisted operator selection framework that can contain any number of operators and be embedded into any CMOEA. We embedded the proposed method into four existing CMOEAs. Experimental studies have demonstrated that the proposed adaptive operator selection method is effective, and the resulting algorithm outperformed nine state-of-the-art CMOEAs.

{
Nevertheless, some issues must be addressed regarding the current study of this paper. According to the experimental results, DRL-assisted CMOEAs perform worse when a CMOP prefers specific operators. Therefore, it is necessary to improve learning efficiency so that fewer computational resources can train a more accurate DRL model to determine the desired operators. Moreover, increasing the maximum number of function evaluations in the experiments may improve the numerical performance of DRL-assisted CMOEAs, which is also true when the proposed method is applied to real-world problems.
}

In the future, the following directions are worth trying:
\begin{itemize}
 \item Some other operators can be embedded to extend this framework to solve other kinds of MOPs. For example, the CSO~\cite{Zhang2018competitive} or other enhanced operators~\cite{Tian2022Integrating} can be embedded to solve large-scale CMOPs~\cite{Tian2022Fast}. {In addition, some advanced learning-based optimizers such as switching particle swarm optimizers~\cite{Li2022ranking-system-based,Zeng2022Dynamic} can be used as actions to enhance performance.}
 \item Advanced neural networks~\cite{Guo2022Evolutionary} can be employed as the DQN to see if they can enhance the performance of the proposed DQL-assisted operator selection method.
 \item {Moreover, hyperparameter adaptation~\cite{Luo2022Position-Transitional} is necessary for future study since there are many hyperparameters that need to be adjusted.}
\end{itemize}

The codes of the methods in this work can be obtained from the authors upon request.

\bibliographystyle{IEEEtran}
\bibliography{Reference}

\begin{thebibliography}{10}
\providecommand{\url}[1]{#1}
\csname url@samestyle\endcsname
\providecommand{\newblock}{\relax}
\providecommand{\bibinfo}[2]{#2}
\providecommand{\BIBentrySTDinterwordspacing}{\spaceskip=0pt\relax}
\providecommand{\BIBentryALTinterwordstretchfactor}{4}
\providecommand{\BIBentryALTinterwordspacing}{\spaceskip=\fontdimen2\font plus
\BIBentryALTinterwordstretchfactor\fontdimen3\font minus
  \fontdimen4\font\relax}
\providecommand{\BIBforeignlanguage}[2]{{%
\expandafter\ifx\csname l@#1\endcsname\relax
\typeout{** WARNING: IEEEtran.bst: No hyphenation pattern has been}%
\typeout{** loaded for the language `#1'. Using the pattern for}%
\typeout{** the default language instead.}%
\else
\language=\csname l@#1\endcsname
\fi
#2}}
\providecommand{\BIBdecl}{\relax}
\BIBdecl

\bibitem{Kumar2021Benchmark-Suite}
A.~Kumar, G.~Wu, M.~Z. Ali, Q.~Luo, R.~Mallipeddi, P.~N. Suganthan, and S.~Das,
  ``A benchmark-suite of real-world constrained multi-objective optimization
  problems and some baseline results,'' \emph{Swarm and Evolutionary
  Computation}, vol.~67, p. 100961, 2021.

\bibitem{Tan2021Evolutionary}
B.~Tan, H.~Ma, Y.~Mei, and M.~Zhang, ``Evolutionary multi-objective
  optimization for web service location allocation problem,'' \emph{IEEE
  Transactions on Services Computing}, vol.~14, no.~2, pp. 458--471, 2021.

\bibitem{Ma2021Shift}
Z.~Ma and Y.~Wang, ``Shift-based penalty for evolutionary constrained
  multiobjective optimization and its application,'' \emph{IEEE Transactions on
  Cybernetics}, pp. 1--13, 2021, doi:{10.1109/TCYB.2021.3069814}.

\bibitem{Tian2020Coevolutionary}
Y.~Tian, T.~Zhang, J.~Xiao, X.~Zhang, and Y.~Jin, ``A coevolutionary framework
  for constrained multiobjective optimization problems,'' \emph{IEEE
  Transactions on Evolutionary Computation}, vol.~25, no.~1, pp. 102--116,
  2021.

\bibitem{Tian2021Balancing}
Y.~Tian, Y.~Zhang, Y.~Su, X.~Zhang, K.~C. Tan, and Y.~Jin, ``Balancing
  objective optimization and constraint satisfaction in constrained
  evolutionary multiobjective optimization,'' \emph{IEEE Transactions on
  Cybernetics}, vol.~52, no.~9, pp. 9559--9572, 2022.

\bibitem{Jiao2022Multiform}
R.~Jiao, B.~Xue, and M.~Zhang, ``A multiform optimization framework for
  constrained multiobjective optimization,'' \emph{IEEE Transactions on
  Cybernetics}, pp. 1--13, 2022, doi:{10.1109/TCYB.2022.3178132}.

\bibitem{Qiao2022An}
K.~Qiao, K.~Yu, B.~Qu, J.~Liang, H.~Song, and C.~Yue, ``An evolutionary
  multitasking optimization framework for constrained multiobjective
  optimization problems,'' \emph{IEEE Transactions on Evolutionary
  Computation}, vol.~26, no.~2, pp. 263--277, 2022.

\bibitem{Ma2021New}
Z.~Ma, Y.~Wang, and W.~Song, ``A new fitness function with two rankings for
  evolutionary constrained multiobjective optimization,'' \emph{IEEE
  Transactions on Systems, Man, and Cybernetics: Systems}, vol.~51, no.~8, pp.
  5005--5016, 2021.

\bibitem{Tian2022Principled}
Y.~Tian, X.~Zhang, C.~He, K.~Tan, and Y.~Jin, ``Principled design of
  translation, scale, and rotation invariant variation operators for
  metaheuristics,'' \emph{Chinese Journal of Electronics}, 07 2022.

\bibitem{Holland1994Adaptation}
J.~Holland, ``Adaptation in natural and artificial systems,'' \emph{An
  Introductory Analysis with Application to Biology, Control and Artificial
  Intelligence}, 01 1994.

\bibitem{Storn1997Differential}
R.~Storn and K.~Price, ``Differential evolution - a simple and efficient
  heuristic for global optimization over continuous spaces,'' \emph{Journal of
  Global Optimization}, vol.~11, pp. 341--359, 01 1997.

\bibitem{Obayyanahatti1995New}
R.~Eberhart and J.~Kennedy, ``A new optimizer using particle swarm theory,'' in
  \emph{MHS'95. Proceedings of the Sixth International Symposium on Micro
  Machine and Human Science}, 1995, pp. 39--43.

\bibitem{Zhang2018competitive}
X.~Zhang, X.~Zheng, R.~Cheng, J.~Qiu, and Y.~Jin, ``A competitive mechanism
  based multi-objective particle swarm optimizer with fast convergence,''
  \emph{Information Sciences}, vol. 427, pp. 63--76, 2018.

\bibitem{Hu2023FCAN-MOPSO}
L.~Hu, Y.~Yang, Z.~Tang, Y.~He, and X.~Luo, ``Fcan-mopso: An improved
  fuzzy-based graph clustering algorithm for complex networks with
  multi-objective particle swarm optimization,'' \emph{IEEE Transactions on
  Fuzzy Systems}, pp. 1--16, 2023.

\bibitem{Hu2020Variational}
L.~Hu, K.~C.~C. Chan, X.~Yuan, and S.~Xiong, ``A variational bayesian framework
  for cluster analysis in a complex network,'' \emph{IEEE Transactions on
  Knowledge and Data Engineering}, vol.~32, no.~11, pp. 2115--2128, 2020.

\bibitem{Wang2019Evolutionary}
C.~Wang, R.~Xu, and X.~Zhang, ``An evolutionary algorithm based on
  multi-operator ensemble for multi-objective optimization,'' in
  \emph{Intelligent Computing Theories and Application}, D.-S. Huang,
  V.~Bevilacqua, and P.~Premaratne, Eds.\hskip 1em plus 0.5em minus 0.4em\relax
  Cham: Springer International Publishing, 2019, pp. 14--24.

\bibitem{Tian2022Deep}
Y.~Tian, X.~Li, H.~Ma, X.~Zhang, K.~C. Tan, and Y.~Jin, ``Deep reinforcement
  learning based adaptive operator selection for evolutionary multi-objective
  optimization,'' \emph{IEEE Transactions on Emerging Topics in Computational
  Intelligence}, pp. 1--14, 2022, doi:{10.1109/TETCI.2022.3146882}.

\bibitem{Dong2022Adaptive}
L.~Dong, Q.~Lin, Y.~Zhou, and J.~Jiang, ``Adaptive operator selection with
  test-and-apply structure for decomposition-based multi-objective
  optimization,'' \emph{Swarm and Evolutionary Computation}, vol.~68, p.
  101013, 2022.

\bibitem{Schneider2021Self-Learning}
S.~Schneider, R.~Khalili, A.~Manzoor, H.~Qarawlus, R.~Schellenberg, H.~Karl,
  and A.~Hecker, ``Self-learning multi-objective service coordination using
  deep reinforcement learning,'' \emph{IEEE Transactions on Network and Service
  Management}, vol.~18, no.~3, pp. 3829--3842, 2021.

\bibitem{Caviglione2021Deep}
L.~Caviglione, M.~Gaggero, M.~Paolucci, and R.~Ronco, ``Deep reinforcement
  learning for multi-objective placement of virtual machines in cloud
  datacenters,'' \emph{Soft Computing}, vol.~25, pp. 1--20, 10 2021.

\bibitem{Liu2022Hybridization}
W.~Liu, R.~Wang, T.~Zhang, K.~Li, W.~Li, and H.~Ishibuchi, ``Hybridization of
  evolutionary algorithm and deep reinforcement learning for multi-objective
  orienteering optimization,'' \emph{IEEE Transactions on Evolutionary
  Computation}, pp. 1--1, 2022, doi:{10.1109/TEVC.2022.3199045}.

\bibitem{Li2022Many-Objective}
Y.~Li, G.~Hao, Y.~Liu, Y.~Yu, Z.~Ni, and Y.~Zhao, ``Many-objective distribution
  network reconfiguration via deep reinforcement learning assisted optimization
  algorithm,'' \emph{IEEE Transactions on Power Delivery}, vol.~37, no.~3, pp.
  2230--2244, 2022.

\bibitem{Zhao2023Hyperheuristic}
F.~Zhao, S.~Di, and L.~Wang, ``A hyperheuristic with q-learning for the
  multiobjective energy-efficient distributed blocking flow shop scheduling
  problem,'' \emph{IEEE Transactions on Cybernetics}, vol.~53, no.~5, pp.
  3337--3350, 2023.

\bibitem{Li2021Deep}
K.~Li, T.~Zhang, and R.~Wang, ``Deep reinforcement learning for multiobjective
  optimization,'' \emph{IEEE Transactions on Cybernetics}, vol.~51, no.~6, pp.
  3103--3114, 2021.

\bibitem{Zhang2022Meta-Learning-Based}
Z.~Zhang, Z.~Wu, H.~Zhang, and J.~Wang, ``Meta-learning-based deep
  reinforcement learning for multiobjective optimization problems,'' \emph{IEEE
  Transactions on Neural Networks and Learning Systems}, pp. 1--14, 2022,
  doi:{10.1109/TNNLS.2022.3148435}.

\bibitem{Fan2019Push}
Z.~Fan, W.~Li, X.~Cai, H.~Li, C.~Wei, Q.~Zhang, K.~Deb, and E.~Goodman, ``Push
  and pull search for solving constrained multi-objective optimization
  problems,'' \emph{Swarm and Evolutionary Computation}, vol.~44, pp. 665--679,
  2019.

\bibitem{Zhu2020Constrained}
Q.~{Zhu}, Q.~{Zhang}, and Q.~{Lin}, ``A constrained multiobjective evolutionary
  algorithm with detect-and-escape strategy,'' \emph{IEEE Transactions on
  Evolutionary Computation}, vol.~24, no.~5, pp. 938--947, 2020.

\bibitem{Alejandro2021Micro-Genetic}
A.~Santiago, B.~Dorronsoro, H.~J. Fraire, and P.~Ruiz, ``Micro-genetic
  algorithm with fuzzy selection of operators for multi-objective optimization:
  $\mu$fame,'' \emph{Swarm and Evolutionary Computation}, vol.~61, p. 100818,
  2021.

\bibitem{Yuan2015An}
Y.~Yuan, H.~Xu, and B.~Wang, ``An experimental investigation of variation
  operators in reference-point based many-objective optimization,'' in
  \emph{Proceedings of the 2015 Annual Conference on Genetic and Evolutionary
  Computation}, ser. GECCO '15.\hskip 1em plus 0.5em minus 0.4em\relax New
  York, NY, USA: Association for Computing Machinery, 2015, p. 775–782.

\bibitem{McClymont2011Markov}
K.~McClymont and E.~C. Keedwell, ``Markov chain hyper-heuristic (mchh): An
  online selective hyper-heuristic for multi-objective continuous problems,''
  in \emph{Proceedings of the 13th Annual Conference on Genetic and
  Evolutionary Computation}, ser. GECCO '11.\hskip 1em plus 0.5em minus
  0.4em\relax New York, NY, USA: Association for Computing Machinery, 2011, p.
  2003–2010.

\bibitem{Lin2022One-to-one}
A.~Lin, P.~Yu, S.~Cheng, and L.~Xing, ``One-to-one ensemble mechanism for
  decomposition-based multi-objective optimization,'' \emph{Swarm and
  Evolutionary Computation}, vol.~68, p. 101007, 2022.

\bibitem{Sutton1998Reinforcement}
R.~Sutton and A.~Barto, ``Reinforcement learning: An introduction,'' \emph{IEEE
  Transactions on Neural Networks}, vol.~9, no.~5, pp. 1054--1054, 1998.

\bibitem{Fayaz2022Machine}
S.~Fayaz, M.~Zaman, and M.~Butt, \emph{Machine Learning: An Introduction to
  Reinforcement Learning}, 07 2022, pp. 1--22.

\bibitem{Watkins1992Q-Learning}
C.~Watkins and P.~Dayan, ``Technical note: Q-learning,'' \emph{Machine
  Learning}, vol.~8, pp. 279--292, 05 1992.

\bibitem{Mnih2015Human-level}
V.~Mnih, K.~Kavukcuoglu, D.~Silver, A.~Rusu, J.~Veness, M.~Bellemare,
  A.~Graves, M.~Riedmiller, A.~Fidjeland, G.~Ostrovski, S.~Petersen,
  C.~Beattie, A.~Sadik, I.~Antonoglou, H.~King, D.~Kumaran, D.~Wierstra,
  S.~Legg, and D.~Hassabis, ``Human-level control through deep reinforcement
  learning,'' \emph{Nature}, vol. 518, pp. 529--33, 02 2015.

\bibitem{Kerschke2019Automated}
P.~Kerschke, H.~H. Hoos, F.~Neumann, and H.~Trautmann, ``{Automated Algorithm
  Selection: Survey and Perspectives},'' \emph{Evolutionary Computation},
  vol.~27, no.~1, pp. 3--45, 03 2019.

\bibitem{Gong2015Adaptive}
W.~Gong, Z.~Cai, and D.~Liang, ``Adaptive ranking mutation operator based
  differential evolution for constrained optimization,'' \emph{IEEE
  Transactions on Cybernetics}, vol.~45, no.~4, pp. 716--727, 2015.

\bibitem{Zhang2008Multiobjective}
Q.~Zhang, A.~Zhou, S.~Zhao, P.~Suganthan, W.~Liu, and S.~Tiwari,
  ``Multiobjective optimization test instances for the cec 2009 special session
  and competition,'' \emph{Mechanical Engineering}, 01 2008.

\bibitem{Fan2020Difficulty}
Z.~Fan, W.~Li, X.~Cai, H.~Li, C.~Wei, Q.~Zhang, K.~Deb, and E.~Goodman,
  ``Difficulty adjustable and scalable constrained multiobjective test problem
  toolkit,'' \emph{Evolutionary Computation}, vol.~28, no.~3, pp. 339--378,
  2020.

\bibitem{Liu2019Handling}
Z.~{Liu} and Y.~{Wang}, ``Handling constrained multiobjective optimization
  problems with constraints in both the decision and objective spaces,''
  \emph{IEEE Transactions on Evolutionary Computation}, vol.~23, no.~5, pp.
  870--884, 2019.

\bibitem{Fan2019An}
Z.~Fan, W.~Li, X.~Cai, H.~Huang, Y.~Fang, Y.~Yugen, J.~Mo, C.~Wei, and
  E.~Goodman, ``An improved epsilon constraint-handling method in moea/d for
  cmops with large infeasible regions,'' \emph{Soft Computing}, vol.~23, 12
  2019.

\bibitem{Ming2021Dual-Population}
M.~Ming, A.~Trivedi, R.~Wang, D.~Srinivasan, and T.~Zhang, ``A
  dual-population-based evolutionary algorithm for constrained multiobjective
  optimization,'' \emph{IEEE Transactions on Evolutionary Computation},
  vol.~25, no.~4, pp. 739--753, 2021.

\bibitem{Liu2021Handling}
Z.-Z. Liu, B.-C. Wang, and K.~Tang, ``Handling constrained multiobjective
  optimization problems via bidirectional coevolution,'' \emph{IEEE
  Transactions on Cybernetics}, vol.~52, no.~10, pp. 10\,163--10\,176, 2022.

\bibitem{Li2014Shift-Based}
M.~{Li}, S.~{Yang}, and X.~{Liu}, ``Shift-based density estimation for
  pareto-based algorithms in many-objective optimization,'' \emph{IEEE
  Transactions on Evolutionary Computation}, vol.~18, no.~3, pp. 348--365,
  2014.

\bibitem{Yu2022Dynamic}
K.~Yu, J.~Liang, B.~Qu, Y.~Luo, and C.~Yue, ``Dynamic selection
  preference-assisted constrained multiobjective differential evolution,''
  \emph{IEEE Transactions on Systems, Man, and Cybernetics: Systems}, vol.~52,
  no.~5, pp. 2954--2965, 2022.

\bibitem{Yuan2022constrained}
J.~Yuan, H.-L. Liu, and Z.~He, ``A constrained multi-objective evolutionary
  algorithm using valuable infeasible solutions,'' \emph{Swarm and Evolutionary
  Computation}, vol.~68, p. 101020, 2022.

\bibitem{Agrawal2000Simulated}
R.~Agrawal, K.~Deb, and R.~Agrawal, ``Simulated binary crossover for continuous
  search space,'' \emph{Complex Systems}, vol.~9, pp. 115--148, 06 2000.

\bibitem{Ishibuchi2019Comparison}
H.~Ishibuchi, R.~Imada, N.~Masuyama, and Y.~Nojima, ``Comparison of
  hypervolume, igd and igd+ from the viewpoint of optimal distributions of
  solutions,'' in \emph{Evolutionary Multi-Criterion Optimization}, 01 2019,
  pp. 332--345.

\bibitem{Ishibuchi2022Difficulties}
H.~Ishibuchi, L.~M. Pang, and K.~Shang, ``Difficulties in fair performance
  comparison of multi-objective evolutionary algorithms [research frontier],''
  \emph{IEEE Computational Intelligence Magazine}, vol.~17, no.~1, pp. 86--101,
  2022.

\bibitem{KEEL}
J.~Alcal{\'a}-Fdez, L.~S{\'a}nchez, S.~Garc{\'i}a, M.~J. del Jesus, S.~Ventura,
  J.~M. Garrell, J.~Otero, C.~Romero, J.~Bacardit, V.~M. Rivas, J.~C.
  Fern{\'a}ndez, and F.~Herrera, ``{KEEL}: {A} software tool to assess
  evolutionary algorithms for data mining problems,'' \emph{Soft Comput.},
  vol.~13, no.~3, pp. 307--318, 2009.

\bibitem{Schott1995Fault}
J.~Schott, ``Fault tolerant design using single and multicriteria genetic
  algorithm optimization,'' p. 203, 05 1995.

\bibitem{Tian2022Integrating}
Y.~Tian, H.~Chen, H.~Ma, X.~Zhang, K.~C. Tan, and Y.~Jin, ``Integrating
  conjugate gradients into evolutionary algorithms for large-scale continuous
  multi-objective optimization,'' \emph{IEEE/CAA Journal of Automatica Sinica},
  vol.~9, no.~10, pp. 1801--1817, 2022.

\bibitem{Tian2022Fast}
Y.~Tian, Y.~Feng, X.~Zhang, and C.~Sun, ``A fast clustering based evolutionary
  algorithm for super-large-scale sparse multi-objective optimization,''
  \emph{IEEE/CAA Journal of Automatica Sinica}, vol.~10, no.~4, pp. 1048--1063,
  2023.

\bibitem{Li2022ranking-system-based}
H.~Li, J.~Li, P.~Wu, Y.~You, and N.~Zeng, ``A ranking-system-based switching
  particle swarm optimizer with dynamic learning strategies,''
  \emph{Neurocomputing}, vol. 494, pp. 356--367, 2022.

\bibitem{Zeng2022Dynamic}
N.~Zeng, Z.~Wang, W.~Liu, H.~Zhang, K.~Hone, and X.~Liu, ``A dynamic
  neighborhood-based switching particle swarm optimization algorithm,''
  \emph{IEEE Transactions on Cybernetics}, vol.~52, no.~9, pp. 9290--9301,
  2022.

\bibitem{Guo2022Evolutionary}
D.~Guo, X.~Wang, K.~Gao, Y.~Jin, J.~Ding, and T.~Chai, ``Evolutionary
  optimization of high-dimensional multiobjective and many-objective expensive
  problems assisted by a dropout neural network,'' \emph{IEEE Transactions on
  Systems, Man, and Cybernetics: Systems}, vol.~52, no.~4, pp. 2084--2097,
  2022.

\bibitem{Luo2022Position-Transitional}
X.~Luo, Y.~Yuan, S.~Chen, N.~Zeng, and Z.~Wang, ``Position-transitional
  particle swarm optimization-incorporated latent factor analysis,'' \emph{IEEE
  Transactions on Knowledge and Data Engineering}, vol.~34, no.~8, pp.
  3958--3970, 2022.

\end{thebibliography}

\begin{IEEEbiography}[{\includegraphics[width=1in,height=1.25in,clip,keepaspectratio]{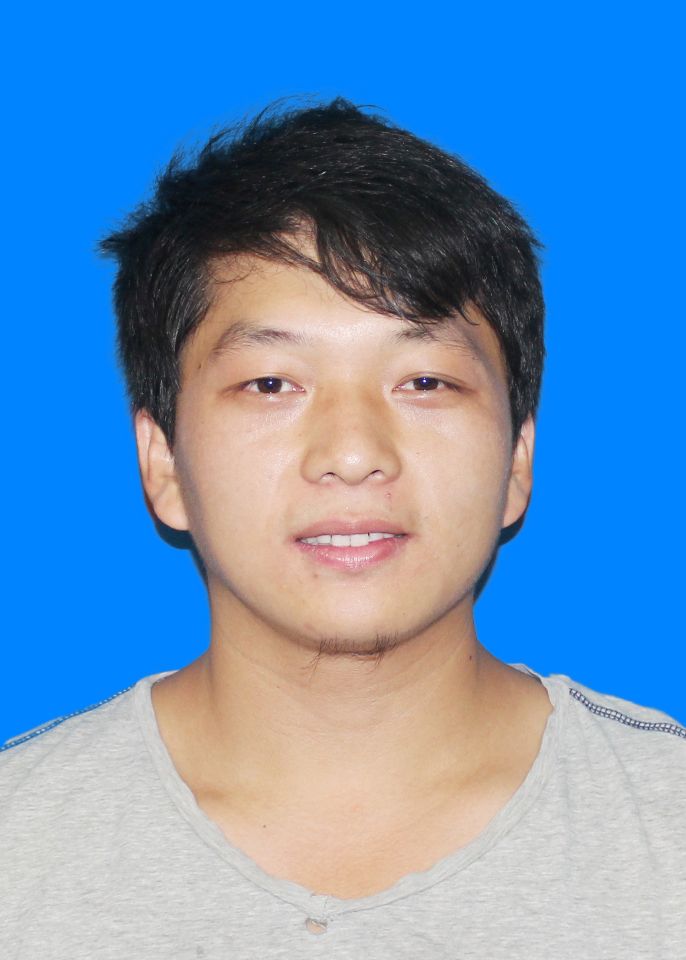}}]{Ming Fei}
	received the B.Sc. degree from China University of Geosciences, Wuhan, China, in 2019. He is currently pursuing the Ph.D degree with the School of Computer, China University of Geosciences, Wuhan, China.
	
	His current research interests include evolutionary multiobjective optimization methods and their applications.
\end{IEEEbiography}

\begin{IEEEbiography}[{\includegraphics[width=1in,height=1.25in,clip,keepaspectratio]{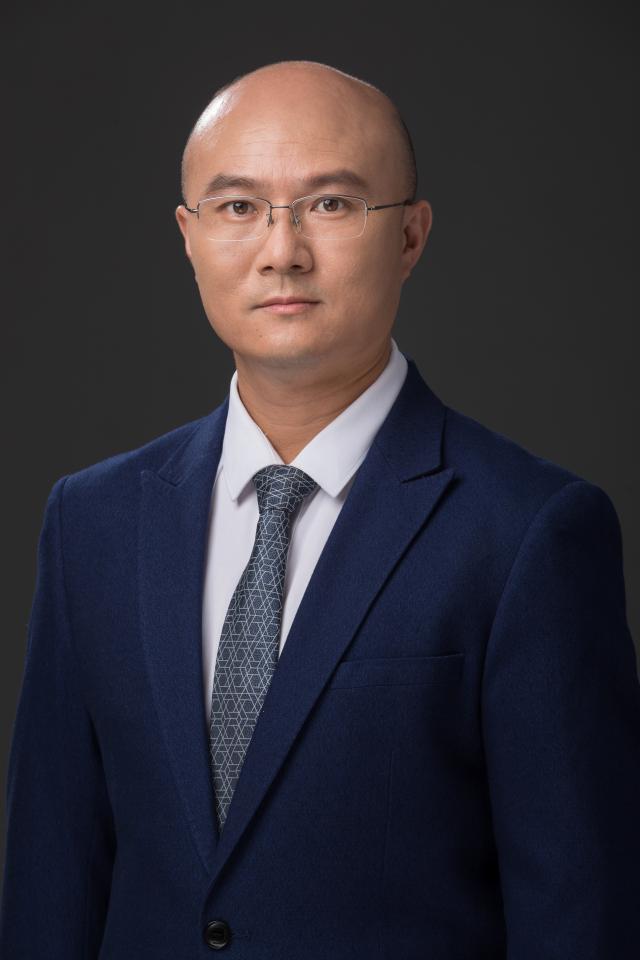}}]{Wenyin Gong (Member, IEEE)}
received the B.Eng., M.Eng., and Ph.D. degrees in computer science from China University of Geosciences, Wuhan, China, in 2004, 2007, and 2010, respectively.

He is currently a Professor with School of Computer Science, China University of Geosciences, Wuhan, China. His research interests include evolutionary algorithms, evolutionary optimization, and their applications. He has published over $80$ research papers in journals and international conferences.

He served as a referee for over $30$ international journals, such as {\sc IEEE Transactions on Evolutionary Computation, IEEE Transactions on Cybernetics, IEEE Transactions on Systems, Man, and Cybernetics: Systems}, \emph{IEEE Computational Intelligence Magazine}, \emph{ACM Transactions on Intelligent Systems and Technology}, \emph{Information Sciences}, \emph{European Journal of Operational Research}, \emph{Applied Soft Computing}, \emph{Journal of Power Sources}, etc. Professor Gong currently serves as Associate Editor of \emph{Expert Systems with Applications}, \emph{International Journal of Bio-Inspired Computation}, \emph{Memetic Computing}, etc.
\end{IEEEbiography}

\begin{IEEEbiography}[{\includegraphics[width=1in,height=1.25in,clip,keepaspectratio]{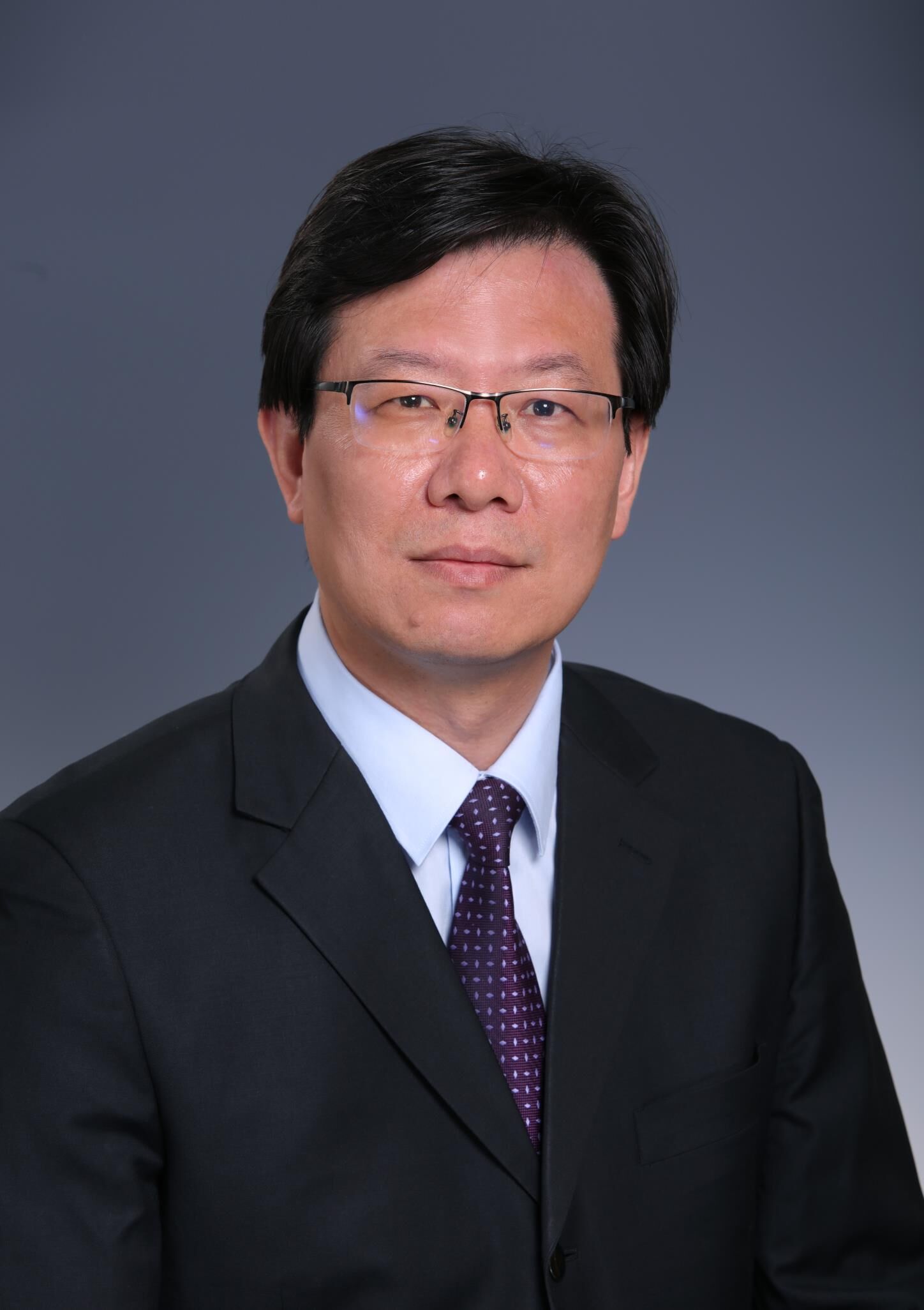}}]{Ling Wang (Member, IEEE)}
received the B.Sc. in automation and Ph.D. in control theory and control engineering from Tsinghua University, Beijing, China, in 1995 and 1999, respectively.

Since 1999, he has been with the Department of Automation, Tsinghua University, where he became a full professor in 2008. His current research interests include intelligent optimization and production scheduling. He has authored five academic books and more than 300 refereed papers.

Professor Wang is a recipient of the National Natural Science Fund for Distinguished Young Scholars of China, the National Natural Science Award (Second Place) in 2014, the Science and Technology Award of Beijing City in 2008, the Natural Science Award (First Place in 2003, and Second Place in 2007) nominated by the Ministry of Education of China. Professor Wang now is the Editor-in-Chief of \emph{International Journal of Automation and Control}, and the Associate Editor of {\sc{IEEE Transactions on Evolutionary Computation}}, \emph{Swarm and Evolutionary Computation}, \emph{Expert Systems with Applications}, etc.

\end{IEEEbiography}

\begin{IEEEbiography}[{\includegraphics[width=1in,height=1.25in,clip,keepaspectratio]{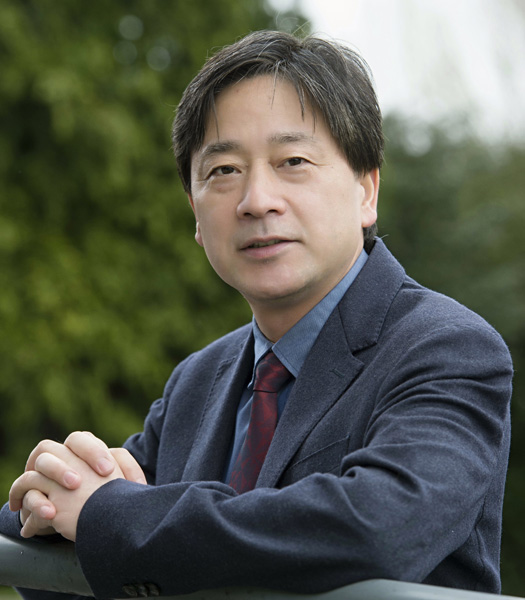}}]{Yaochu Jin (Fellow, IEEE)}
received the B.Sc., M.Sc., and Ph.D. degrees in automatic control from Zhejiang University, Hangzhou, China, in 1988, 1991, and 1996, respectively, and the Dr.-Ing. degree in neuroinformatics from Ruhr University Bochum, Bochum, Germany, in 2001.

He is currently an Alexander von Humboldt Professor of artificial intelligence with the Faculty of Technology, Bielefeld University, Bielefeld, Germany, endowed by the German Federal Ministry of Education and Research. He is also the Distinguished Chair and a Professor of computational intelligence with the Department of Computer Science, University of Surrey, Guildford, U.K. His main research interests include multiobjective and data-driven evolutionary optimization, evolutionary multiobjective learning, trustworthy AI, and evolutionary developmental AI.

Dr. Jin is a member of Academia Europaea. He was named by the Web of Science as ``a Highly Cited Researcher'' from 2019 to 2022 consecutively.

\end{IEEEbiography}

\end{document}